\documentclass{article}

\usepackage[preprint,nonatbib]{neurips_2024}

\usepackage[utf8]{inputenc} 
\usepackage[T1]{fontenc}    
\usepackage{hyperref}       
\usepackage{url}            
\usepackage{booktabs}       
\usepackage{amsfonts}       
\usepackage{nicefrac}       
\usepackage{microtype}      
\usepackage{xcolor}         
\usepackage{amsmath}
\usepackage{amsthm}
\usepackage{graphicx}
\usepackage{subfig}
\usepackage{enumitem}

\usepackage{algorithmic}
\usepackage{algorithm}

\newif\ifannote

\ifannote
    
    \newcommand{\anncomment}[3]{
        {\color{#1}\colorbox{#1}{\bfseries\sffamily\tiny\textcolor{white}{#2}}
        $\triangleright$ \em #3 $\triangleleft$}
    }
\else
    
    \newcommand{\anncomment}[3]{}
\fi

\newtheorem{theorem}{Theorem}[section]

\newtheorem{remark}[theorem]{Remark}

\title{Overcoming the Challenges of Batch \\
 Normalization in Federated Learning}

\author{%
    Rachid Guerraoui \\ EPFL \And Rafael Pinot \\ Sorbonne Université \And Geovani Rizk\thanks{Correspondence to: Geovani Rizk <geovani.rizk@epfl.ch>} \\ EPFL \And John Stephan \\ EPFL \And François Taiani \\ Univ Rennes, Inria, CNRS, IRISA
}

\begin{document}

\newcommand{\fbn}{FBN}
\newcommand{\bn}{BatchNorm}

\maketitle

\begin{abstract}
  Batch normalization has proven to be a very beneficial mechanism to accelerate the training and improve the accuracy of deep neural networks in centralized environments. Yet, the scheme faces significant challenges in federated learning, especially under high data heterogeneity. Essentially, the main challenges arise from external covariate shifts and inconsistent statistics across clients.
  We introduce in this paper Federated \bn{} (FBN), a novel scheme that restores the benefits of batch normalization in federated learning.
  Essentially, FBN ensures that the batch normalization during training is consistent with what would be achieved in a centralized execution, hence preserving the distribution of the data, and providing running statistics that accurately approximate the global statistics.
  \fbn{} thereby reduces the external covariate shift and matches the evaluation performance of the centralized setting.
  We also show that, with a slight increase in complexity, 
  we can robustify FBN to mitigate erroneous statistics and potentially adversarial attacks. 
\end{abstract}

\section{Introduction}
\label{sec:intro}

Federated learning (FL) has emerged as a prominent machine learning paradigm~\cite{mcmahan17fedavg}, owing to its ability to train models without requiring centralized data storage~\cite{kairouz2021advances}. In FL, the training data is held by multiple devices (referred to as \emph{clients}), and the training process is orchestrated by a central server, which coordinates with the clients to train a model on the union of their datasets. Crucially, clients do not share their raw data with the server, thereby substantially reducing the lack of data sovereignty commonly associated with traditional centralized learning methods. In short, this distributed approach ensures that sensitive information remains on local devices, enhancing data ownership and security. Although FL is likely to become a standard in modern machine learning, it remains poorly understood/used in many respects e.g.,~\cite{kairouz2021advances}. In particular, some of the powerful techniques that have been 
developed to boost the performance of centralized learning do not have their analogue in distributed environments. 

\vspace{2pt}\textbf{Batch Normalization.} \bn{} is a prominent example of such techniques. 
This popular scheme in centralized deep learning is widely used when training convolutional or feed-forward neural networks~\cite{BatchNormIoffe2015,BatchNormOptimTsipras2018,DeepConvBatchNormSimonyanZ2014}. \bn{} stabilizes and accelerates the learning by normalizing the inputs to each layer of the network using the mean and variance of the previous layer's outputs. During the training phase, \bn{} performs two essential tasks. First, it reduces the \emph{internal covariate shift} of the network, ensuring that the input to each layer has a consistent distribution throughout the entire training procedure~\cite{BatchNormIoffe2015}. Second, \bn{} maintains \emph{running statistics}, such as the running mean and running variance of each layer, over the batches of data being used during training. These running statistics provide a smoothed estimate of the data distribution of each layer, which happens to be crucial for the evaluation phase. In evaluation mode, \bn{} normalizes the test inputs using the running mean and variance obtained during the training phase, instead of computing batch-specific statistics. This ensures that the model's behavior is consistent across inputs (and independent of the composition of the test mini-batch), resulting in more stable and reliable predictions~\cite{BatchNormIoffe2015}.

\vspace{2pt}\textbf{Centralized vs Federated Implementations.} Implementing \bn{} in a centralized setting does not raise any serious challenges. Indeed, as the data is stored on a single machine, the mini-batch being used at any given point during the training procedure constitutes a good approximation of the entire training set. As a result, both the normalization of layers and the computation of running statistics are easy to perform and relatively consistent across batches. In FL, however, each client holds its own local dataset which may not be a faithful representation of the entire population. This hinders the direct application of \bn{} both during the training and the \textcolor{black}{evaluation phase~\cite{wang2023batch}}. Essentially, the heterogeneity between clients' local datasets leads to inconsistencies between local statistics that can affect the overall training procedure, potentially slowing down convergence and reducing model accuracy. Furthermore, during the evaluation phase, most existing implementations simply average the running means and variances of the clients \cite{beutel2020flower, roth2022nvidia}. While this is a reasonable strategy when clients have identical datasets, when local datasets are heterogeneous, this creates an important bias in the computation of the running variance that can destroy the predictive performance of the model.
Arguably, the need to deal with the statistical disparities of the clients and to ensure consistent model performance requires to rethink the way \bn{} is implemented in FL. We tackle this challenge by introducing an improved version of \bn{} for FL, called Federated \bn{} (\fbn{}) which, while rather simple, is quite powerful for it matches the performance of centralized \bn{} learning. Our contributions can be summarized as follows.

\subsection{Contributions}

\begin{enumerate}[leftmargin=10pt]
    \item We first explain why straightforward implementations of BatchNorm in distributed environments fail to match the performance of centralized methods. The rationale behind this phenomenon is twofold. On the one hand, since the clients normalize the layers' inputs using only their individual statistics, they essentially never have access to the global information from other clients. By considering the special case of a 1-layer neural network with local Gaussian distributions, we show that this lack of global information makes the data less separable, hence making the learning procedure much harder to perform. On the other hand, the averaging of the running variance during the evaluation phase creates an important bias that further reduces the performance of the distributed implementation compared to the centralized one. 

    \item We then introduce Federated BatchNorm (\fbn{}), an improved version of \bn{} that aims to align the normalization processes in FL with that of a centralized setting. The rationale behind \fbn{} is to have the server send the running statistics of previous rounds to the clients to serve as a good approximation of the global statistics of the network. However, as mentioned above, these statistics can be biased due to the \textcolor{black}{\emph{external covariate shift}~\cite{BatchNormIoffe2015}}. We circumvent this drawback by unbiasing the running variance, thus ensuring that the aggregated statistics are computed correctly. Essentially, \fbn{} provides a dual solution: (i) it ensures that the batch normalization \emph{during training} is consistent, hence preserving the separability of the data, and (ii) it provides running statistics that accurately approximate the global statistics, thereby reducing the external covariate shift and allowing to match the evaluation performance of the centralized setting. 
    We empirically evaluate \fbn{} on the CIFAR-10~\cite{cifar} classification task under several levels of data heterogeneity, and show that it matches the performance of centralized \bn{} learning.  

    \item Finally, in light of the recent literature on the vulnerability of FL methods to Byzantine clients~\cite{guerraoui2023byzantine}, we evaluate the robustness of \fbn{} to this type of threats. We show that the robustness of \fbn{} can easily be enhanced through the use of robust aggregation techniques such as \emph{coordinate-wise median} or \emph{coordinate-wise trimmed mean}~\cite{yin2018byzantine}. We empirically evaluate these defense mechanisms, and show that our algorithm defends better against this threat model than the traditional implementation of \bn{} in FL.  
\end{enumerate}

    
\subsection{Related Work}\label{sec_related_work}

Several recent works have attempted to align the behavior of BatchNorm in FL with that of the centralized setting. Notably, \cite{wang2023batch} proposes a method where local statistics are computed and transmitted to the server during the forward pass. The server then aggregates these statistics and returns the result of the aggregation, enabling the clients to normalize their data using the global distribution. Clients must however wait, at each layer of the model, for the server's response before proceeding, as the output of one layer impacts the input of subsequent ones. This results in communication overhead and latency, rendering the method very time-consuming. \fbn{} does not suffer from the same limitation as it essentially requires the same amount of communication as a standard FL scheme.  

Alternatively, \cite{zhong2023making} introduces FixBN, which operates in two phases. In the first phase, FixBN functions exactly like BatchNorm to reach a point where the gradients do not vary too much. In the second phase, FixBN stops using batch-specific statistics, and uses the running mean and variance aggregated by the server during the first phase. While FixBN significantly enhances the final performance of BatchNorm in FL, it has two important limitations. First, the model's performance is sub-optimal during the first phase, requiring prior knowledge of the total number of learning steps, which is not necessary for BatchNorm in centralized settings. Second, the running mean and variance are biased, and the statistics communicated by the clients do not allow the server to correct this bias. Our approach circumvents these limitations, as it is agnostic to the number of training steps and sends sufficient information to the server to compute unbiased estimates of the global statistics. 


Another line of research has focused on leveraging alternatives to \bn{} developed in centralized settings. For instance,~\cite{hsieh2020non} uses Group Normalization (introduced in ~\cite{wu2018group}) which is independent of batch size and computes statistics on groups of channels within the models.
Alternatively,~\cite{du2022rethinking} claims that the covariate shift cannot be avoided in FL and suggests to use Layer Normalization (first introduced in~\cite{ba2016layer}) instead of \bn{}. This solution renders the normalization process batch size-independent and effective for recurrent neural networks. \cite{li2021fedbn} also proposes to only use \bn{} in personalized FL settings so that each client can retain their own statistics and parameters. Finally,\cite{zhuang2024fedwon} proposes a method that removes the \bn{} component and rather reparameterizes convolution layers with scaled weight standardization. Although these approaches represent significant progress in overcoming the challenges associated with \bn{} in FL, they are mainly motivated by the alleged impracticality of accurately computing distributed batch norm statistics. Our work shows that \bn{} can be used efficiently in a distributed environment, without having to resort to complex methods and without impacting the communication efficiency of the training algorithm.








\section{Background on \bn{} in the Centralized Setting}\label{sec:BNCentralized}
Consider a standard classification task with an input space $\mathcal{X}$, and an output space $\mathcal{Y}$. We focus on a class of neural networks of the form $F_\theta(x) := \phi^{(N)}_{\theta}\circ \cdots \circ  \phi^{(1)}_{\theta}(x)$, 
where $\theta \in \mathbb{R}^p$ represents the trainable parameters of the model and $\phi^{(i)}_\theta$ is called the $i$-th layer of $F_\theta$. 
In a centralized setting, we assume the machine has access to a training set $S$ and attempts to minimize an \emph{empirical loss} $$\mathcal{L}(\theta) := \frac{1}{\vert S \vert} \sum_{(x,y)\in S} \ell\left(F_{\theta}(x),y \right),$$ where $\ell$ is a point-wise loss function (e.g., cross entropy).
The most common approach to this minimization problem in machine learning is to run a mini-batch stochastic gradient descent (SGD) algorithm.
SGD iteratively computes gradients on mini-batches of data sampled from the training set $S$.
The driving principle of \bn{} is to use the specificity of these mini-batch computations to stabilize and accelerate the training.
In short, given a mini-batch of data from $\{x_1, \dots, x_b\}$, \bn{} normalizes the inputs of each layer $\phi^{(j)}$ within $F_\theta$ during the forward pass, i.e., when computing $\{F_{\theta}(x_1), ..., F_{\theta}(x_b)\}$.
This scheme also updates a running mean and a running variance for each layer $\phi^{(j)}$ to keep track of the ongoing state of the algorithm, and to use during the evaluation phase.
Since \bn{} is applied successively to a set of vectors of varying sizes (i.e., the inputs of each layer of $F_\theta$), we present below the algorithm on $K$ arbitrary values $\{u_1, ..., u_K \} \in \mathbb{R}^K$.
When dealing with vectors of higher dimension, the same computations simply apply coordinate-wise.




\vspace{2pt}\textbf{Summary of the method.} \bn{} takes as input a batch of $K$ values $\{u_1, \dots, u_K\} \in \mathbb{R}^K$, a momentum parameter $\beta \in (0, 1)$, a numerical stability parameter $\epsilon > 0$, as well as a running mean $\bar{u}_{\text{run}}$ and variance $\hat{\sigma}_{\text{run}}^2$. Given these inputs, the method consists in normalizing the batch according to~\eqref{normalize_func} by using their empirical mean $\bar{u} = \frac{1}{K} \sum_{k = 1}^K u_k$ and (biased) variance $\hat{\sigma}^2 = \frac{1}{K} \sum_{k = 1}^K \left(u_k - \bar{u}\right)^2$.
In~\eqref{normalize_func}, the parameters $\alpha$ and $\zeta$ are used to reshape the normalized inputs, and are learnable parameters for the neural network at hand. 
Upon normalizing the inputs, \bn{} then updates both the running mean and variance, using the momentum parameter $\beta$ and the mean and variance computed above, as per line 5 of Algorithm~\ref{BN}. 
Finally, \bn{} outputs the normalized inputs, and the updated running mean and variance. The full procedure is summarized in Algorithm~\ref{BN}. 

\begin{algorithm}[tb]
   \caption{\bn{}}
   \label{BN}
\begin{algorithmic}[1]
    \STATE {\bfseries Input:} Batch of values $\{u_1, \dots, u_K\} \in \mathbb{R}^K$, momentum $\beta \in (0,1)$, numerical stability parameter $\varepsilon>0$, running mean $\bar{u}_{\text{run}}$, running variance $\hat{\sigma}_{\text{run}}^2$.
    \STATE Compute current mean $\bar{u} = \frac{1}{K} \sum_{k = 1}^K u_k$ and variance $\hat{\sigma}^2 = \frac{1}{K} \sum_{k = 1}^K \left(u_k - \bar{u}\right)^2$
    \STATE Normalize data \footnotemark\vspace{-1.2em}
    
    \begin{equation}
        (\tilde{u}_1, \dots, \tilde{u}_K) = \left(\frac{u_1 - \bar{u}}{\sqrt{\hat{\sigma}^2+\varepsilon}}, \dots, \frac{u_K - \bar{u}}{\sqrt{\hat{\sigma}^2+\varepsilon}} \right) * \alpha + \zeta
        \label{normalize_func}
    \end{equation}
    \STATE Update running mean $\bar{u}_{\text{run}} \leftarrow (1-\beta) \times \bar{u}_\text{run} + \beta \times \bar{u}$
    \STATE Update running variance $\left(\hat{\sigma}_{\text{run}}\right)^2\leftarrow (1-\beta) \times \hat{\sigma}_{\text{run}}^2 + \beta \times \frac{K}{K-1} \times \hat{\sigma}^2$
    \STATE {\bfseries Output:}  $\left\{(\tilde{u}_1, \dots, \tilde{u}_K), \bar{u}_{\text{run}}, \hat{\sigma}_{\text{run}}^2\right\}$
\end{algorithmic}
\end{algorithm}

\section{Bottlenecks of \bn{} in Distributed Environments}\label{sec:BNFailsinFL}
In FL, a group of $n$ clients communicating through a central server aim to train a global model over the collection of their data.
The clients hold local datasets $S_1, \dots, S_n$, which are assumed to be of the same size for the sake of simplicity.
The objective of the training procedure is to find the parameter that minimizes the average of the local loss functions \footnote{When the local datasets $S_i$ have different sizes, the formulation changes slightly and involves weights.}
$$\frac{1}{n} \sum_{i = 1}^n\mathcal{L}_i(\theta), \text{ where } \mathcal{L}_i(\theta) := \frac{1}{\vert S_i \vert} \sum_{(x,y)\in S_i} \ell\left(F_{\theta}(x),y \right), ~~ \forall i \in \{1 ,..., n\}.$$
We would like to leverage the benefits of \bn{} in stabilizing the training on a central machine, and transfer them to a distributed setting comprising multiple clients.
\textcolor{black}{To this aim, we need to adapt 
\bn{}
to the federated setting. Simply using the straightforward implementations of BatchNorm contained in off-the-shelf distributed libraries such as~\cite{beutel2020flower, roth2022nvidia} 
would cause each client to execute Algorithm~\ref{BN} on their local batches at every local training step, before sending their running statistics to the server at every communication step.~\footnote{We distinguish between local training and server communication because communicating with the server need not happen after every local training step, as is the case of many FL algorithms\textcolor{black}{~\cite{mcmahan17fedavg, karimireddy2020scaffold}}}}
Upon receiving the running statistics from every client, the server \emph{averages them}, i.e., computes $$\bar{u}_{\text{run}} = \frac{1}{n} \sum_{i = 1}^n \bar{u}_{\text{run}}^{(i)} \text{ {} {} and {} {}
  } \hat{\sigma}_{\text{run}}^2 = \frac{1}{n} \sum_{i = 1}^n \left( \hat{\sigma}_{\text{run}}^{(i)} \right )^2,$$
where $\bar{u}_{\text{run}}^{(i)}$ and $(\hat{\sigma}_{\text{run}}^{(i)})^{2}$ respectively denote the running mean and variance sent by client $i \in \{1,...,n\}$ at the current communication round.
This protocol (hereafter referred to as \emph{Naive BatchNorm}) suffers from two important limitations: (i) it does not handle the external covariate shift due to local normalization, and (ii) it does not accurately estimate the \textit{global} running statistics of the clients.

\begin{figure}[t]
    \centering
    \subfloat[Centralized setting]{
    \label{fig_separability_centr}
    \includegraphics[width=0.4\textwidth]{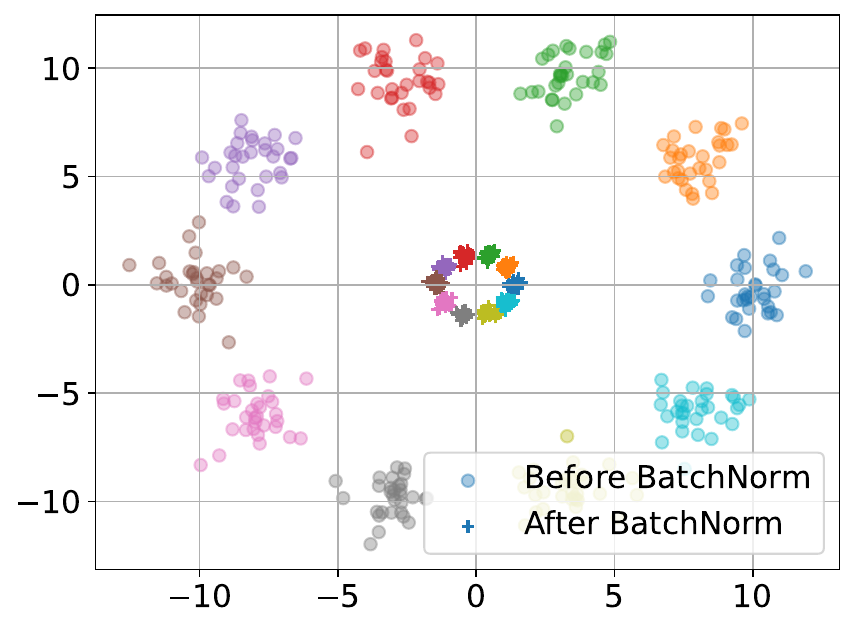}}
    \subfloat[Federated setting (Naive)]{\includegraphics[width=0.4\textwidth]{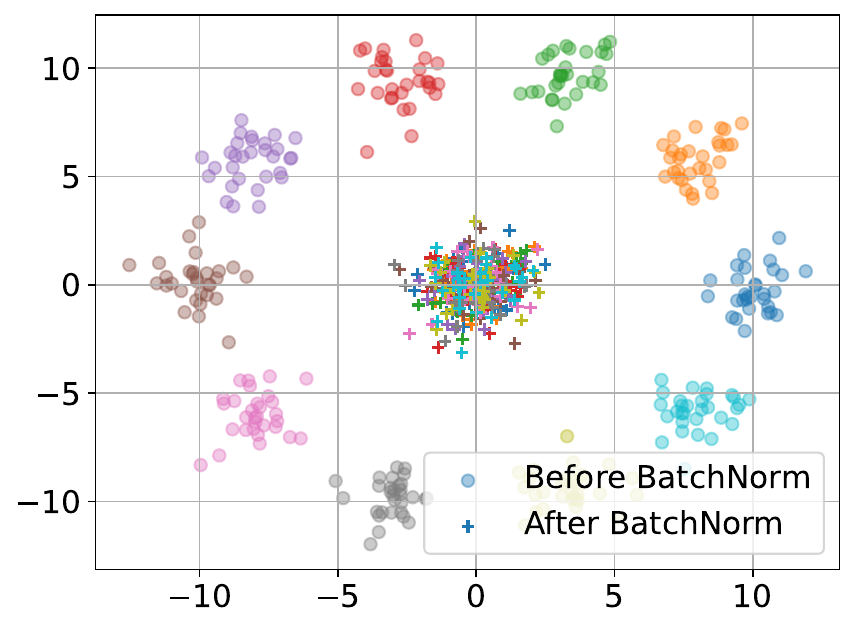}}
    \caption{Points normalized with \bn{} in a centralized setting (a) and  with Naive \bn{} in a federated setting (b). In the latter, the points are distributed heterogeneously: each client has one different class that is represented by one color. In a centralized setting the data is properly normalized while the normalization fails with Naive \bn{} 
    ultimately making the points non-separable.
    }
\label{data_points_batchnorms}
\end{figure}

\vspace{2pt}\textbf{Local normalization breaks data separability.} To better understand why \bn{} suffers from a covariate shift in FL, consider the following simple example with $\mathcal{X}=\mathbb{R}^2$ and $\mathcal{Y}=\{1,...,10\}$.
We consider an FL setting with $n=10$ clients that each 
hold a dataset $S_i$ of the form $$S_i = \left \{ \left(x^{(i)}_1,i \right), ..., \left(x^{(i)}_m,i \right) \right \}, \text{ where } x^{(i)}_1, ..., x^{(i)}_m \underset{i.i.d}{\sim} \mathcal{N}(\mu_i, s^2 I_2),$$
with $\mu_i \in \mathbb{R}^2$, $s^2 > 0$, $I_2$ is the $2$-dimensional identity matrix, and $i$ is the label of all datapoints in $S_i$.
In other words, we consider a highly heterogeneous scenario, where every client $i$ has exactly one unique class in their dataset (coded as $i$, which is different from the others), and where the features of each class follow a distinct normal distribution.
Furthermore, we focus on models having a single layer ($N=1$), which translates to only observing the impact of \bn{} on the data batches being sampled from the clients. Finally, we assume that the clients sample their entire dataset to execute \bn{}. We illustrate this setting through a numerical experiment in Figure~\ref{data_points_batchnorms}, with $\mu_i = 10\left(\cos{\left(2\pi \nicefrac{(i-1)}{10}\right)}, \sin{\left(2\pi \nicefrac{(i-1)}{10}\right)}\right)$, $s^2 =1$, and $m=30$. In this example, each class is linearly separable before applying any normalization.

In a centralized execution, i.e., if we place all the clients' data on a single machine before applying \bn{}, the normalization does not change the linear separability of the data. This phenomenon is illustrated in Figure~\ref{data_points_batchnorms} (left), where the datapoints before (``$\bullet$'' markers) and after (``$+$'' markers) normalization essentially continue to form the same pattern after being rescaled.
By contrast, in the federated case, normalization~\eqref{normalize_func} is applied locally by each client on their data. Then, as illustrated in Figure~\ref{data_points_batchnorms} (right), the data is no longer separable after local normalization.
This discrepancy in the normalization outputs of both settings can be traced back to the distribution of data on the clients.
In the centralized setting, the empirical mean $\bar{u}$ is close to the point $(0,0)$ and barely affects the numerator of the normalization function~\eqref{normalize_func}. Every group of points of the same class thus remains more or less in the same region of the feature space before being scaled down in magnitude by the denominator $\sqrt{\hat{\sigma}^2+\epsilon}$.
By contrast, in the federated case, every client computes a different empirical mean $\bar{u}^{(i)}$ and shifts all its points towards $(0,0)$ (when subtracting the mean, see~\eqref{normalize_func}). As a result, the different classes of the clients overlap in $\mathbb{R}^2$ post normalization.
We believe that this loss of separability is a primary factor contributing to the sub-optimal performance of \bn{} in FL.



\vspace{2pt}\textbf{Biased running statistics during evaluation.} Recall that in naive implementations of \bn{} in FL, the server simply averages the running statistics it receives from the clients. 
While this computation correctly estimates the global running mean $\bar{u}_{\text{run}}$, it introduces bias when used to estimate the global running variance $\hat{\sigma}_{\text{run}}^2$, as it violates the law of total variance. During the evaluation phase, this discrepancy may further cause the performance of the FL scheme to diverge from that of the centralized one, and this phenomenon can be amplified by data heterogeneity.

To circumvent both these limitations and match the performance of the centralized setting, every client should ideally have access to the global statistics as if they were centrally computed on the union of all the clients' batches.
This can be achieved during the forward pass if each client communicates their statistics of every layer of $F_\theta$ to the server, and wait to receive the aggregated statistics before normalizing their local batch.
However, as explained in Section~\ref{sec_related_work}, the clients must wait for the server's response at each layer of the model before proceeding.
While this may work if the number of layers $N$ is small enough, when considering deep neural networks, this approach is not a practical solution due to its high communication costs, \textcolor{black}{as shown in~\cite{wang2023batch}}.
\section{Federated BatchNorm (\fbn{})}\label{sec_fed_setting}


\begin{algorithm}[t]
   \caption{Federated BatchNorm (FBN)%
   }
   \label{FBN}
\begin{algorithmic}[1]
    \STATE {\bfseries Input for each client $i$:} Batch $\{u_1^{(i)}, \dots, u_K^{(i)}\} \in \mathbb{R}^K$, momentum $\beta \in (0,1)$, number of clients $n \in \mathbb{N}$, numerical stability parameter $\varepsilon>0$, (previous) shared running statistics $\bar{u}_{\text{run}}$ and $\hat{\sigma}_{\text{run}}^2$.
    \vspace{5pt}
    \FOR{\text{each client $i$ \textbf{in parallel}}}
        \STATE Normalize data using previous running statistics
        \begin{align}
            (\tilde{u}_1^{(i)}, \dots, \tilde{u}_K^{(i)}) = \left( \frac{u_1^{(i)} - \textcolor{purple}{\bar{u}_{\text{run}}}}{\sqrt{\textcolor{purple}{\hat{\sigma}_{\text{run}}^2}+\varepsilon}}, \dots, \frac{u_K^{(i)} - \textcolor{purple}{\bar{u}_{\text{run}}}}{\sqrt{\textcolor{purple}{\hat{\sigma}_{\text{run}}^2}+\varepsilon}}\right) * \alpha + \zeta
        \end{align}
        \STATE Compute current mean $\bar{u}^{(i)} = \frac{1}{K} \sum_{k = 1}^K u_k^{(i)}$ and variance $\left(\hat{\sigma}^{(i)}\right)^2 = \frac{1}{K} \sum_{k = 1}^K \left(u_k^{(i)} - \bar{u}^{(i)}\right)^2$
        \STATE Update local running mean $\bar{u}_{\mathrm{run}}^{(i)} \leftarrow (1-\beta) \times \bar{u}_\text{run} + \beta \times \bar{u}^{(i)}$ 
        \vspace{5pt}
        \STATE Update local running variance $\left(\hat{\sigma}_{\mathrm{run}}^{(i)}\right)^2 \leftarrow (1-\beta) \times \hat{\sigma}_{\text{run}}^2 + \beta \times \frac{K\textcolor{purple}{n}}{K\textcolor{purple}{n}-1} \times \left(\hat{\sigma}^{(i)}\right)^2$
        \STATE Send $\bar{u}_\text{run}^{(i)}$ and  $\left(\hat{\sigma}_{\text{run}}^{(i)}\right)^2$ to Server
    \ENDFOR
    \vspace{5pt}
    \STATE Server updates running mean $\bar{u}_{\mathrm{run}} = \frac{1}{n} \sum_{i=1}^n \bar{u}_{\mathrm{run}}^{(i)}$ \label{alg:runningmean}
    \STATE Server updates running var. $\hat{\sigma}_{\mathrm{run}}^2 = \frac{1}{n} \sum_{i=1}^n \left(\hat{\sigma}_{\mathrm{run}}^{(i)}\right)^2 + \textcolor{purple}{\frac{Kn}{(Kn-1)\beta} \times \frac{1}{n}\sum_{i = 1}^n \left(\bar{u}^{(i)}_{\mathrm{run}} - \bar{u}_{\mathrm{run}}\right)^2}$ \label{alg:runningvar}
    \STATE Server sends back the updates shared running statistics $\bar{u}_\text{run}$ and $\left(\hat{\sigma}_{\text{run}}\right)^2$ to all clients.
    \vspace{5pt}
    \STATE {\bfseries Each client $i$ outputs:}  $\left(\tilde{u}_1^{(i)}, \dots, \tilde{u}_K^{(i)}\right)$
\end{algorithmic}
\end{algorithm}

In this paper, we propose \emph{Federated BatchNorm} (\fbn{}), an improved implementation of BatchNorm that aims to align the normalization process in FL with that of a centralized setting. Our method is based on the observation that, when running an iterative FL scheme, if the global data distribution of the current rounds is similar to that of the previous rounds, then the running statistics can be used to approximate the current global means and variances.

\vspace{2pt}\textbf{\fbn{} explained.} The core idea behind our solution is to replace the batch-specific statistics by \emph{shared} running statistics when normalizing the batch of every client.
Specifically, we ensure that whenever \fbn{} is used, each client has access to the same running statistics $\bar{u}_{\text{run}}$ and $\hat{\sigma}_{\text{run}}^2$, and that  these are being used to normalize the local batches of inputs (see line 2 in Algorithm~\ref{FBN}). This is intended to avoid external covariate shifts.
In fact, since the running statistics are the same at every client, using them to normalize the inputs of each layer results in the same normalization function across the clients. However, as explained previously, the running variance can be biased when dealing with heterogeneous data.
Hence, the update of the running statistics on the clients' side, as well as their aggregation at the server, must be performed carefully to compensate the bias introduced by heterogeneity. 
Essentially, our goal is to update the running variance as if it was computed in a centralized setting, i.e., applying \bn{} on the union of the clients' inputs.
To do this, following the law of total variance, a term removing the bias in the averaging of the local running variances must be added as follows (see also line 10 in Algorithm~\ref{FBN}):
$$\hat{\sigma}_{\mathrm{run}}^2 = \frac{1}{n} \sum_{i=1}^n \left(\hat{\sigma}_{\mathrm{run}}^{(i)}\right)^2 + \textcolor{purple}{\frac{Kn}{(Kn-1)\beta} \times \frac{1}{n}\sum_{i = 1}^n \left(\bar{x}^{(i)}_{\mathrm{run}} - \bar{x}_{\mathrm{run}}\right)^2}.$$
This correction ensures that both the shared running variance and running mean used by the clients to normalize their batches with \fbn{} are exactly the same as those they would have obtained in a centralized setting (see Appendix~\ref{app_fbn_iterative} for more details). The full scheme is summarized in Algorithm~\ref{FBN}. 

\begin{figure}[ht!]
    \centering
    \subfloat[FBN (ours)]{\hspace{-15pt}\includegraphics[width=0.35\textwidth]{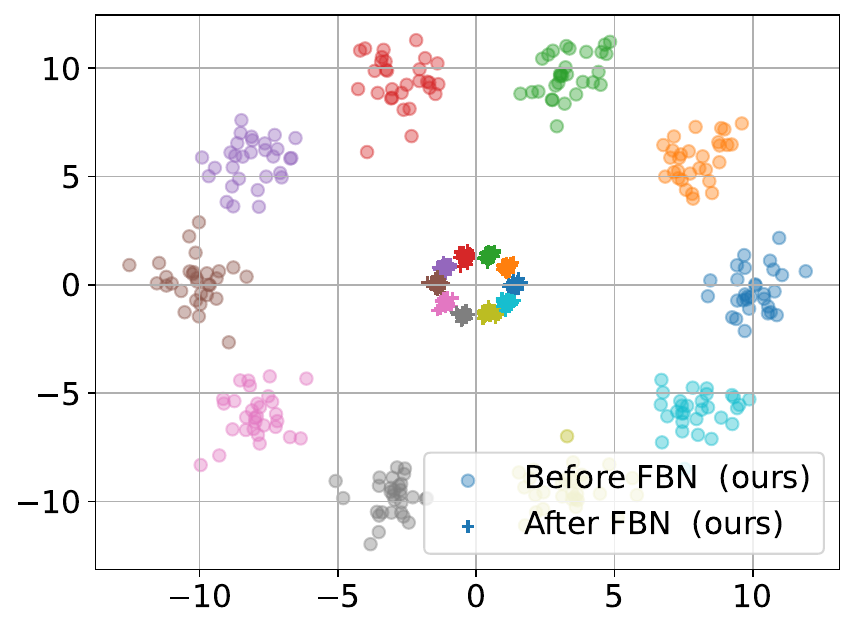}\label{data_points_batchnorms2:FBN}}
    \subfloat[FixBN]{\includegraphics[width=0.35\textwidth]{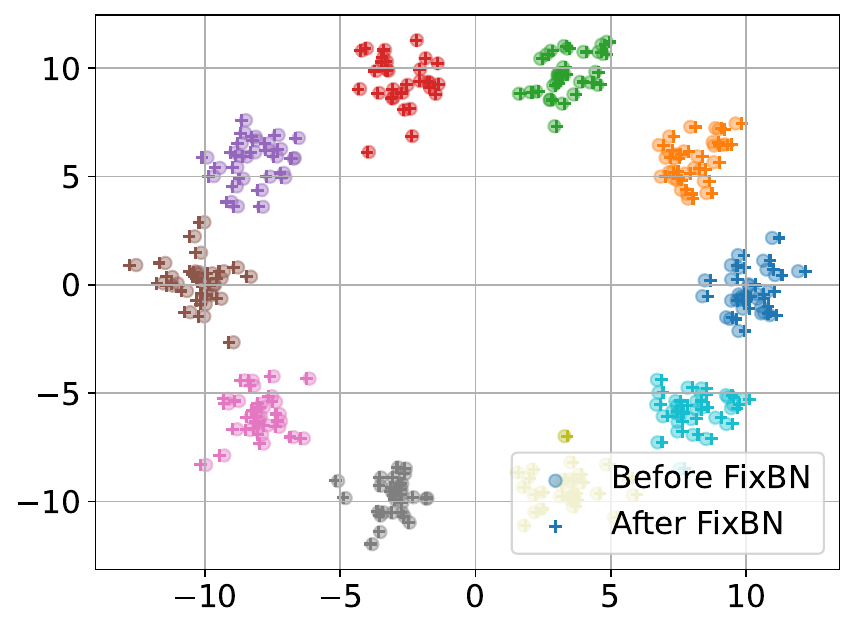}\label{data_points_batchnorms2:FixBN}}
    \subfloat[Error vs centralized case]{\includegraphics[width=0.34\textwidth]{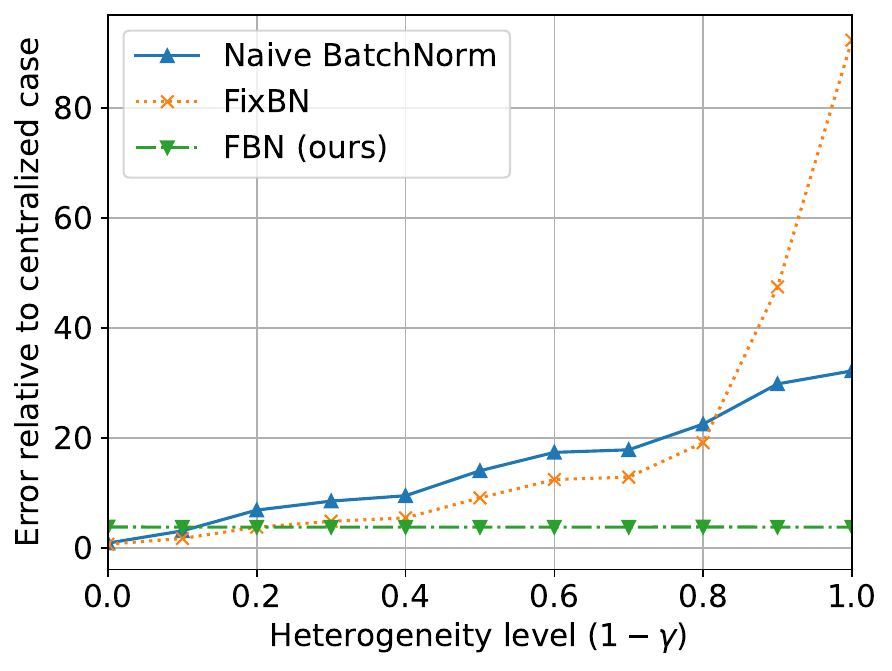}\label{data_points_batchnorms2:error}}
    \caption{(a) and (b) show the points normalized by FBN and FixBN, respectively, in extreme heterogeneity. (c) shows the averaged distance between the normalized points using $\{$FBN, FixBN, Naive BatchNorm$\}$ in a federated setting and the normalized points using \bn{} in the centralized setting for different heterogeneity regimes. For all schemes, we set $\beta = 0.1$. 
    }
\label{data_points_batchnorms2}
\end{figure}

\vspace{2pt}\textbf{\fbn{} preserves data separability.} To 
understand how \fbn{} impacts separability, 
we revisit the example presented in Section~\ref{sec:BNFailsinFL}.
Using the same dataset, we perform $100$ successive rounds of normalization using \fbn{}, where in each round, every client samples a batch of $30$ data points from their respective normal distribution $\mathcal{N}(\mu_i, s^2 I_2)$.
The resulting normalization, shown in Figure~\ref{data_points_batchnorms2:FBN} at iteration $100$, is nearly identical to that obtained in a centralized setting (Figure~\ref{fig_separability_centr}). In particular, it preserves data separability, in contrast to a Naive \bn{} implementation. 
To better understand the interplay between separability and external covariate shift, we run the same experiment with FixBN~\cite{zhong2023making}, the best-known adaptation of \bn{} to FL from the literature (Figure~\ref{data_points_batchnorms2:FixBN}). 
Although FixBN also preserves data separability, it fails to correctly normalize the data due to the bias in the running variance.
Because external covariate shifts arise from data heterogeneity, we show in Figure~\ref{data_points_batchnorms2:error} how each approach (\fbn{}, FixBN, and Naive \bn{}) compares to centralized normalization for different levels of heterogeneity (described in detail below).
For each heterogeneity level (see `Modeling heterogeneity' below for more details on computing each level),
Under four heterogeneity  Figure~\ref{data_points_batchnorms2:error} plots the normalization error of every algorithm, measured as the Euclidean distance between the normalized points and those returned in a centralized setting, averaged over all $100$ rounds.
\fbn{} remains consistently close to that of the centralized case across all levels of heterogeneity, whereas the error of FixBN and Naive \bn{} increases with heterogeneity.

\vspace{2pt}\textbf{Evaluation of \fbn{} on CIFAR-10.} To validate the performance of \fbn{} in FL, we conduct experiments using the classical Distributed Stochastic Gradient Descent (DSGD) algorithm~\cite{bertsekas2015parallel} on CIFAR-10~\cite{cifar}. We refer the interested reader to Appendix~\ref{app:algoDSGD} for more details on this algorithm. 
To assess the accuracy of the model trained using our method, we also implement DSGD with FixBN~\cite{zhong2023making}, and DSGD with Naive \bn{}, and compare to a centralized setting.

\vspace{2pt}\textbf{Modeling heterogeneity.} We consider a distributed setting with $n = 10$ clients and varying levels of data heterogeneity that we model using the notion of $\gamma$-similarity~\cite{karimireddy2020scaffold}. 
Essentially, $\gamma \in [0,1]$ measures the degree of similarity between the clients' datasets.
As $\gamma$ decreases, the clients' datasets become more heterogeneous. To enforce $\gamma$-similarity, we split CIFAR-10 into 2 parts: a homogeneous sub-dataset $D_{\mathit{Ho}}$ of size $\gamma \times len(dataset)$ = $\gamma \times 50,000$, and a heterogeneous sub-dataset $D_{\mathit{He}}$ of size $(1-\gamma) \times 50,000$.
$D_{\mathit{He}}$ is sorted by increasing labels (i.e., the images with label $0$ come first, then label $1$, etc.), and divided sequentially into $n$ chunks $C_1, ..., C_{n}$. Then, each client $i\in \{1,..., 10\}$ samples a fraction $\gamma$ of its dataset uniformly at random from the homogeneous dataset $D_{Ho}$ (without replacement) and adds its corresponding heterogeneous chunk $C_i$ to their dataset.
The clients' datasets are then disjoint and equal in size. When $\gamma = 1$, we say that the clients have iid datasets, while we refer to $\gamma = 0$ as the extreme niid setting.

\begin{remark}
    We choose to evaluate our method on DSGD rather than some other FL algorithm, as the comparison between DSGD and the centralized setting (i.e., SGD) is straightforward. In each round of DSGD, every client samples a mini-batch and computes a local gradient. 
    To have a meaningful comparison with the centralized setting, we simulate a centralized execution by merging in each round the mini-batches of the different clients on which we compute the global gradient.
\end{remark}

\begin{figure}[ht!]
    \centering
    \subfloat[Final acc. vs. Heterogeneity]{\hspace{-15pt}\includegraphics[width=0.34\textwidth]{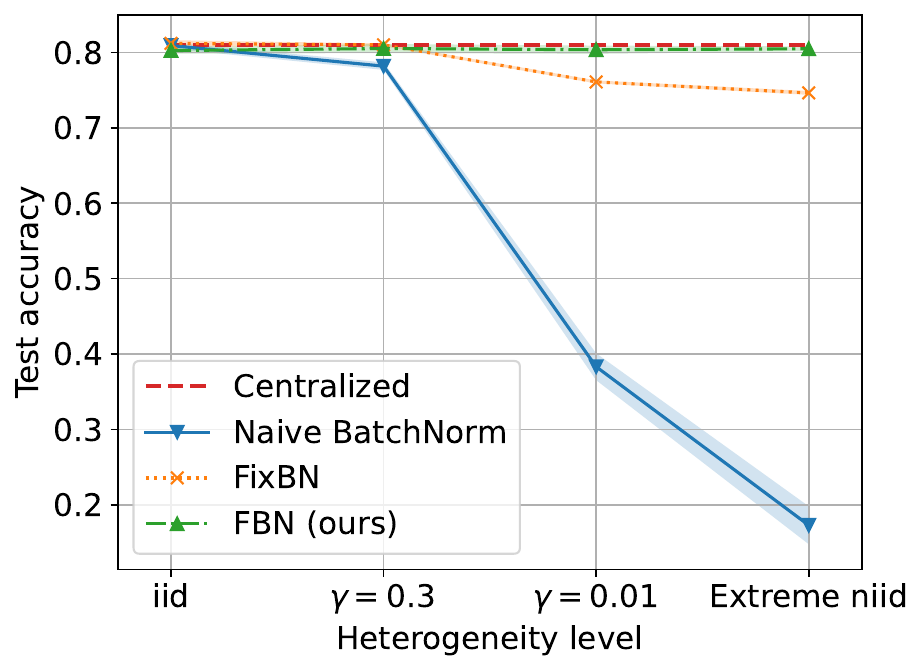}
    \label{fig_comparison_centralized:final}}
    \subfloat[$\gamma=0.01$]{\includegraphics[width=0.34\textwidth]{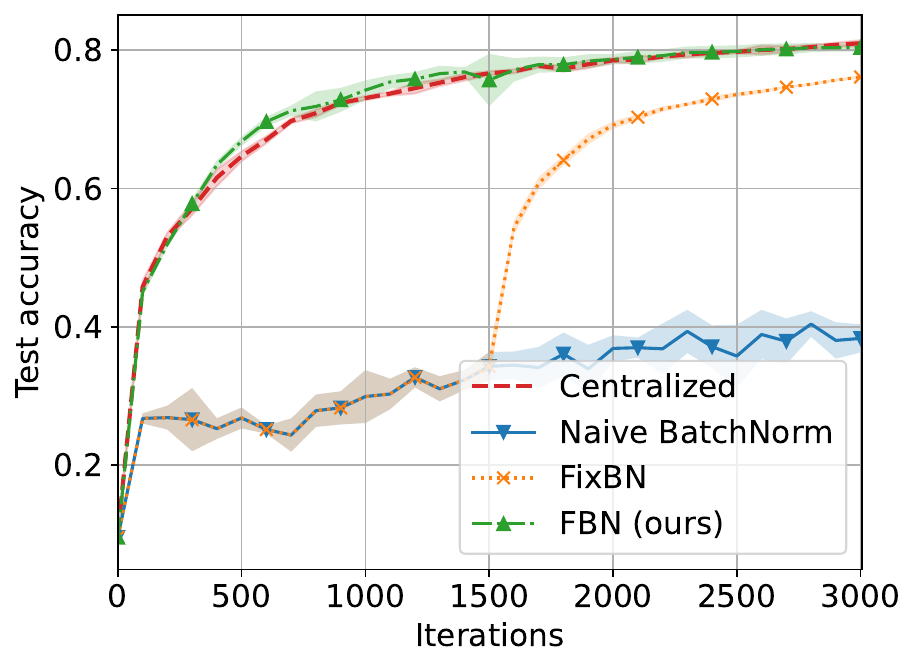}
    \label{fig_comparison_centralized:0_01}}
    \subfloat[$\gamma = 0$]{\includegraphics[width=0.34\textwidth]{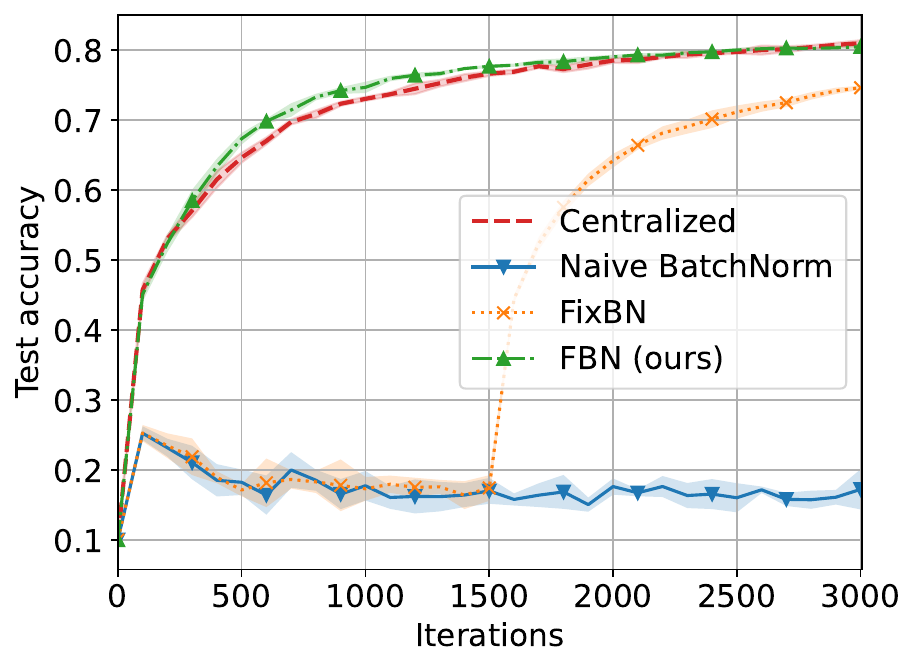}
    \label{fig_comparison_centralized:0}}
    \caption{Final accuracy of DSGD using $\{$\fbn{}, FixBN, Naive \bn{}$\}$ vs heterogeneity (a), evolution of test accuracy for $\gamma = 0.01$ (b) and in extreme heterogeneity ($\gamma=0$) (c).}
    \label{fig_comparison_centralized}
\end{figure}

\vspace{2pt}\textbf{Results.}  Figure~\ref{fig_comparison_centralized:final} plots the final test accuracy reached by the different algorithms under varying heterogeneity levels, while Figures~\ref{fig_comparison_centralized:0_01} and~\ref{fig_comparison_centralized:0} show the evolution of the learning when $\gamma=0.01$ and $\gamma=0$, respectively.
When data heterogeneity is low (i.e., $\gamma \geq 0.3$), the different algorithms perform similarly.
As the setting becomes more heterogeneous, the improvement induced by \fbn{} becomes more apparent.
Indeed, when $\gamma = 0.01$, \bn{} loses 40 points in accuracy to reach 40\%, while the accuracy of FixBN drops only slightly, and that of \fbn{} remains unchanged.
In the extreme niid setting, the performance of Naive \bn{} collapses, delivering a best-case accuracy close to 17\%.
FixBN also exhibits poorer performance, with its maximum accuracy decreasing to 75\%.
Contrastingly, \fbn{} remains unaffected by the increase in heterogeneity, and matches 
the maximal accuracy of 80\% achieved in the centralized setting.
These aforementioned observations are also conveyed in Figures~\ref{fig_comparison_centralized:0_01} and~\ref{fig_comparison_centralized:0}
Moreover, while we observe that FixBN reaches a satisfactory final accuracy under extreme heterogeneity, this algorithm struggles in the first 1500 iterations of the learning (right plot), which correspond to the first phase of the algorithm. Indeed, as mentioned in Section~\ref{sec_related_work}, FixBN operates exactly like \bn{} during its first phase, explaining its sub-optimal performance. In its second phase, FixBN steadily improves in performance and eventually achieves an accuracy of 75\%.
Contrarily, our algorithm does not require prior knowledge of the number of iterations, and almost exactly matches the performance of centralized learning throughout the entire execution in both heterogeneity regimes.

\vspace{-5pt}
\section{On the Robustness of Federated \bn{}}
\label{sec:byz}
When implementing machine learning algorithms in a distributed environment, it is paramount to deal with malfunctioning clients, often referred to as Byzantine clients~\cite{guerraoui2023byzantine}.
These threats encompass a range of adversarial scenarios, ranging from simple errors due to software bugs or hardware issues to more severe issues like data poisoning~\cite{allen2020byzantine, mahloujifar19a}.
The Byzantine threat may also refer to adversarial clients solely aiming to disrupt the learning and preventing it from converging to a satisfactory model. While these issues have been addressed in the context of exchanging gradients or model parameters~\cite{krum, yin2018byzantine, farhadkhani2022byzantine, karimireddy2022byzantinerobust, farhadkhani2023robust, gorbunov2023variance, allouah2023fixing} in distributed learning algorithms (e.g., DSGD, FedAvg), they remain unexplored for distributed batch normalization. In our setting, the presence of Byzantine clients translates to local statistics sent by some clients being incorrect, whether due to intent or inadvertent errors. Averaging all clients' means and variances (as suggested in lines \ref{alg:runningmean}-\ref{alg:runningvar} of Algorithm~\ref{FBN}) is not robust against arbitrary perturbations in these statistics, which could severely impact the training procedure, especially if these faulty statistics are maliciously crafted. This section highlights the inherent vulnerabilities of distributed batch norm implementations in this context. We demonstrate that robust \textit{gradient aggregation} mechanisms proposed in Byzantine ML~\cite{yin2018byzantine, allouah2023fixing} can be effectively applied to make \fbn{} robust to various types of faults.

\vspace{2pt}\textbf{Experimental setup.}
Let $f < \nicefrac{n}{2}$ be the number of Byzantine clients whose identity is a-priori unknown to the server. The remaining $n - f$  clients are correct and follow exactly the prescribed protocol. They are referred to as \textit{honest}. We employ the same ML model used in Section~\ref{sec_fed_setting} in a distributed system of $n = 10$ clients among which $f = 3$ clients are Byzantine. As in the previous experiments, we consider several degrees of heterogeneity in the clients' datasets using $\gamma$-similarity. 
Furthermore, we make the adversarial clients execute three state-of-the-art attacks from the Byzantine ML literature, namely \emph{Sign-flipping} (SF)~\cite{allen2020byzantine}, \emph{Fall of Empires} (FOE)~\cite{xie2020fall}, and \emph{A Little Is Enough} (ALIE)~\cite{baruch2019alittle}. To defend against the attacks, instead of averaging, the server aggregates the clients' statistics using two state-of-the-art robust schemes, namely \emph{coordinate-wise median} and \emph{trimmed mean}~\cite{yin2018byzantine}, pre-processed with a layer of nearest-neighbour mixing (NNM)~\cite{allouah2023fixing}. We test \fbn{} and Naive \bn{} in three scenarios: (a) \textit{No attack} where all clients are honest (i.e., $f = 0$), (b) \textit{No defense} where $f = 3$ but the server does not defend against the attacks (i.e., the server averages the clients' local statistics), (c) \textit{defense} where the server aggregates the statistics using the robust median (or trimmed mean). In this section, we present our results for $f = 3$ adversarial clients executing the SF attack in two heterogeneity settings: $\gamma = 0.01$ and extreme niid ($\gamma = 0$), and defer the remaining results to Appendix~\ref{app_byz_setting}.



\begin{figure*}[ht!]
    \centering
     \subfloat[$\gamma = 0.01$]{\includegraphics[width=0.49\textwidth]{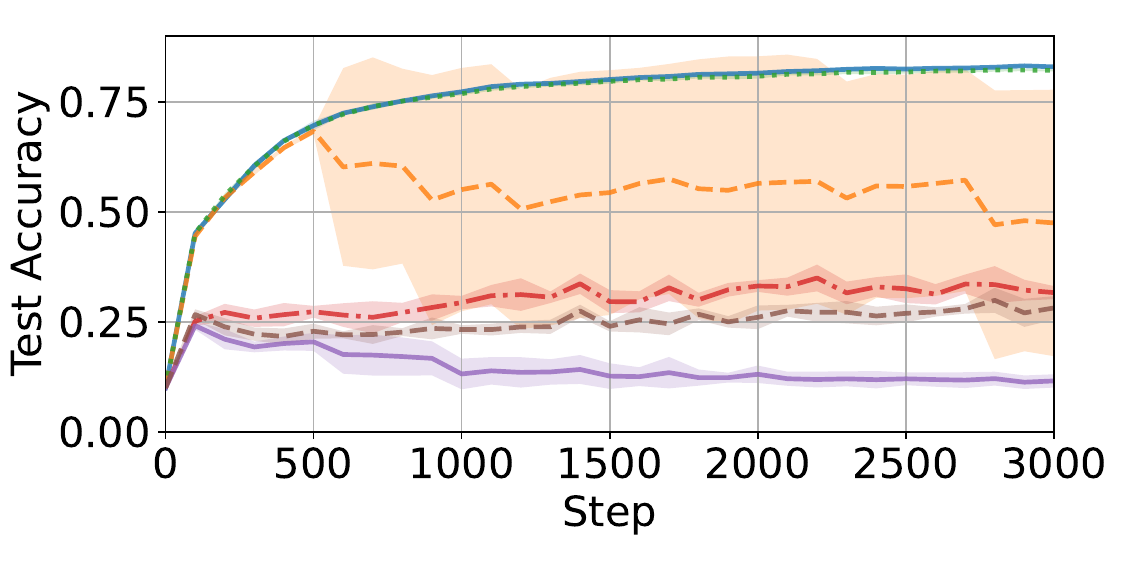}}
    \subfloat[extreme heterogeneity]{\includegraphics[width=0.49\textwidth]{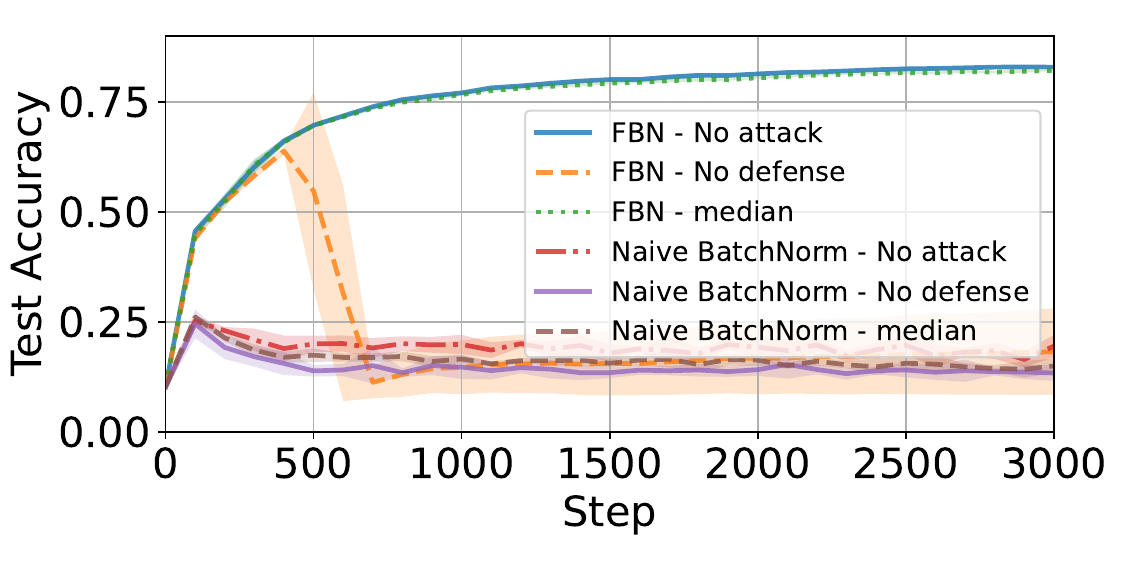}}
    \caption{Performance of \fbn{} vs. Naive \bn{} on CIFAR-10 with adversarial clients in two heterogeneity settings.
    Out of $n=10$ clients, we consider $f = 3$ adversarial clients executing \textbf{SF}~\cite{allen2020byzantine}. Naive \bn{} is strongly impacted by the attack, even when protected through the median~\cite{yin2018byzantine} aggregation. By contrast, \fbn{} (ours) successfully defeats the attack when using the same protection.}
\label{fig_byz_main}
\end{figure*}

\vspace{2pt}\textbf{Empirical results.} In Figure~\ref{fig_byz_main}, we showcase the performance of \fbn{} compared to Naive \bn{}, when $3$ adversarial clients use SF to attack the batch norm statistics sent to the server. Clearly, \textit{not defending} against SF significantly deteriorates the learning for both algorithms.
Indeed, in both heterogeneity settings, Naive \bn{} (purple) is completely unable to learn, stagnating at an accuracy close to 10\%.
Contrastingly, \fbn{} (orange) shows a better learning dynamic where it is able to learn in the beginning reaching an accuracy close to 70\%.
Similarly to Naive \bn{}, not defending against the SF attack proves also to be detrimental for \fbn{}, as its performance quickly degrades especially under extreme heterogeneity.

However, when the server \textit{defends} against SF using the median~\cite{yin2018byzantine}, the performance of both algorithms improves, with a significant edge to \fbn{} nonetheless.
Indeed, when $\gamma = 0.01$, `Naive \bn{} - median' (brown) yields similar accuracies to Naive \bn{} without attack (red), but the baseline algorithm itself is not that performant in this heterogeneity regime, achieving a maximal accuracy of 30\% throughout the learning.
This can also be seen in extreme heterogeneity, where both `Naive \bn{} - No attack' and `Naive \bn{} - median' portray equally poor performance.
On the contrary, \fbn{} combined with the robust median aggregation (green) successfully defends against SF in both heterogeneity settings, and matches almost exactly the performance of \fbn{} without attack (blue).
This illustrates the superior performance of our algorithm in terms of robustness compared to Naive \bn{}, and also highlights the importance of correctly estimating the bias when computing the aggregate statistics at the server.
Furthermore, this analysis clearly showcases the serious vulnerabilities of the distributed batch norm in terms of robustness to faulty statistics.
In fact, a simple attack like SF, where the adversarial clients only modify the mean by sending its opposite, suffices to completely destroy the learning in the absence of a proper defense mechanism at the server.
These observations are also conveyed, in an even stronger fashion, when the adversarial clients execute stronger attacks such as FOE and ALIE.
Indeed, against these attacks, the discrepancy between Naive \bn{} - No attack and Naive \bn{} - median is more significant, while \fbn{} combined with the median completely matches the performance of the algorithm under no attack (see for example Figure~\ref{fig_byz_app_ALIE_median} in Appendix~\ref{app_byz_setting}).

\section{Conclusion}
We have presented a novel and principled approach to apply BatchNorm to FL algorithms.
Our approach, \fbn{}, outperforms existing BatchNorm mechanisms, matching the results of a centralized setting. It is further able to defend against documented adversarial attacks when combined with robust aggregation techniques.  The main idea of our scheme is to replace batch-specific computations by running statistics to avoid blowing up the communication complexity of the algorithm. Interestingly, even-though our method does not use batch-specific information, it is both accurate and efficient. This observation opens an intriguing question on whether normalization needs to be batch-specific in the first place. We believe that further exploring novel normalization algorithms in FL that eschew batches entirely is therefore worth investigating.


\bibliography{references}
\bibliographystyle{abbrv}


\newpage
\appendix
\section{Federated BatchNorm for iterative algorithms}\label{app_fbn_iterative}

In this section we explain how Federated Batch Normalization presented in Algorithm~\ref{FBN} computes the exact same running mean and running variance that would be computed in the centralize case with BatchNorm. To do so, let us consider a given round $t\geq 1$ of a learning algorithm and consider that each client $i \in \{1, ..., n\}$ has a batch of values $\{u^{(i)}_{1,t}, \dots,u^{(i)}_{K,t}\}$ to normalize and we define the \emph{local empirical mean} $\bar{u}_t^{(i)}$ and the \emph{local empirical variance} $\left(\hat{\sigma}_t^{(i)}\right)^2$ as follows

\begin{align*}
    \bar{u}_t^{(i)} = \frac{1}{K} \sum_{k = 1}^K u^{(i)}_{k,t} \ \ \ \ \ \text{ and } \ \ \ \ \ \left(\hat{\sigma}_t^{(i)}\right) = \frac{1}{K} \sum_{k = 1}^K \left(u^{(i)}_{k,t} - \bar{u}_t^{(i)}\right).
\end{align*}

In FBN, each client $i$ updates its running mean $\bar{u}_{\mathrm{run}, t}^{(i)}$ and running variance $\left(\hat{\sigma}_{\mathrm{run}, t}^{(i)}\right)^2$ as follows:

\begin{align*}
    \begin{cases}
    ~~~\bar{u}^{(i)}_{\mathrm{run}, t} &=~ (1-\beta) \times ~~ \bar{u}_{\mathrm{run}, t-1}~ + \beta \times \bar{u}^{(i)}_{t} \\ 
    \left(\hat{\sigma}_{\mathrm{run}, t}^{(i)}\right)^2 &=~ (1-\beta) \times \left(\hat{\sigma}_{\mathrm{run}, t-1}\right)^2 + \beta \times \frac{Kn}{Kn-1} \times \hat{\sigma}^{(i)}_{t}
    \end{cases}
\end{align*}

with $\bar{u}_{\mathrm{run}, 0} = 0$ and $\hat{\sigma}_{\mathrm{run},0}^2 = 1$ by convention.

In the centralized case, i.e., where BatchNorm is applied on the union of all the batches (see Algorithm~\ref{BN}), we compute the \emph{global empirical mean} $\bar{u}_t$ and the \emph{global empirical variance} $\hat{\sigma}_t^2$ as follows


\begin{align*}
    \bar{u}_t = \frac{1}{Kn} \sum_{i = 1}^n \sum_{k = 1}^K u^{(i)}_{k,t} \ \ \ \ \ \text{ and } \ \ \ \ \ \hat{\sigma}_t^2 = \frac{1}{Kn} \sum_{i = 1}^n \sum_{k = 1}^K \left(u^{(i)}_{k,t} - \bar{u}_t\right).
\end{align*}

and update the running means $\bar{u}_{\mathrm{run}, t}$ and running variance $\left(\hat{\sigma}_{\mathrm{run}, t}\right)^2$ as follows:

\begin{align*}
\begin{cases}
    ~~~\bar{u}_{\mathrm{run}, t} &=~ (1-\beta) \times ~~ \bar{u}_{\mathrm{run}, t-1}~ + ~\beta \times \bar{u}_{t} \\ 
    \hat{\sigma}_{\mathrm{run}, t}^2 &=~ (1-\beta) \times \hat{\sigma}_{\mathrm{run}, t-1 }^2 + ~\beta  \frac{Kn}{Kn-1}\times  \hat{\sigma}_{t}.
\end{cases}
\end{align*} 

with $\bar{u}_{\mathrm{run}, 0} = 0$ and $\hat{\sigma}_{\mathrm{run},0}^2 = 1$ by convention.



One can express the running means $\bar{u}_{\mathrm{run}, t}$ computed in the centralized setting as follows
\begin{align*}
    \bar{u}_{\mathrm{run}, t} &= (1-\beta) \times \bar{u}_{\mathrm{run}, t-1} + \beta \times \bar{u}_{t} \\
    &=  (1-\beta) \times \bar{u}_{\mathrm{run}, t-1} + \beta \times \frac{1}{Kn} \sum_{i = 1}^n \sum_{k = 1}^K u^{(i)}_{k,t} \\
    &=  (1-\beta) \times \bar{u}_{\mathrm{run}, t-1} + \beta \times \frac{1}{n} \sum_{i = 1}^n \frac{1}{K}\sum_{k = 1}^K u^{(i)}_{k,t} \\
    &=  (1-\beta) \times \bar{u}_{\mathrm{run}, t-1} + \beta \times \frac{1}{n} \sum_{i = 1}^n \bar{u}^{(i)}_{t} \\
    &=  \frac{1}{n} \sum_{i = 1}^n \left[(1-\beta) \times \bar{u}_{\mathrm{run}, t-1}]\right] + \beta \times \frac{1}{n} \sum_{i = 1}^n \bar{u}^{(i)}_{t}\\
    &=  \frac{1}{n} \sum_{i = 1}^n \left[(1-\beta) \times \bar{u}_{\mathrm{run}, t-1} + \beta \times \bar{u}^{(i)}_{t}\right] \\
    &=  \frac{1}{n} \sum_{i = 1}^n \bar{u}_{\mathrm{run}, t}^{(i)} \tag{RM}\label{run_mean}
\end{align*}

Similarly, one can express the running variance $\left(\hat{\sigma}_{\mathrm{run}, t}\right)^2$ computed in the centralized setting as follows

\begin{align*}
    \hat{\sigma}_{\mathrm{run}, t}^2 &= (1-\beta) \times \hat{\sigma}_{\mathrm{run}, t-1 }^2 + ~\beta  \frac{Kn}{Kn-1}\times  \hat{\sigma}_{t} \\
    &= (1-\beta) \times \hat{\sigma}_{\mathrm{run}, t-1 }^2 + \beta  \frac{Kn}{Kn-1}\times \left(\frac{1}{n} \sum_{i=1}^n \hat{\sigma}_t^{(i)} + \frac{1}{n} \sum_{i=1}^n \left(\bar{u}^{(i)}_{t} - \bar{u}_{t}\right)^2\right)\\
    &= (1-\beta) \times \hat{\sigma}_{\mathrm{run}, t-1 }^2 + \beta  \frac{Kn}{Kn-1}\times \frac{1}{n} \sum_{i=1}^n \hat{\sigma}_t^{(i)} + \beta  \frac{Kn}{Kn-1}\times\frac{1}{n} \sum_{i=1}^n \left(\bar{u}^{(i)}_{t} - \bar{u}_{t}\right)^2 \\
    &= \frac{1}{n} \sum_{i=1}^n \left[(1-\beta) \times \hat{\sigma}_{\mathrm{run}, t-1 }^2 + \beta  \frac{Kn}{Kn-1}\times \hat{\sigma}_t^{(i)}\right] + \beta  \frac{Kn}{Kn-1}\times\frac{1}{n} \sum_{i=1}^n \left(\bar{u}^{(i)}_{t} - \bar{u}_{t}\right)^2 \\
    &= \frac{1}{n} \sum_{i=1}^n \left(\hat{\sigma}_{\mathrm{run}, t}^{(i)}\right)^2 + \beta  \frac{Kn}{Kn-1}\times\frac{1}{n} \sum_{i=1}^n \left(\bar{u}^{(i)}_{t} - \bar{u}_{t}\right)^2 \\
    &= \frac{1}{n} \sum_{i=1}^n \left(\hat{\sigma}_{\mathrm{run}, t}^{(i)}\right)^2 + \beta  \frac{Kn}{Kn-1}\times\frac{1}{n\beta^2} \sum_{i=1}^n \left(\beta \bar{u}^{(i)}_{t} - \beta \bar{u}_{t}\right)^2 \\
    &= \frac{1}{n} \sum_{i=1}^n \left(\hat{\sigma}_{\mathrm{run}, t}^{(i)}\right)^2 +  \frac{Kn}{Kn-1}\times\frac{1}{n\beta} \sum_{i=1}^n \left(\left(1-\beta\right)\bar{u}_{\mathrm{run}, t-1} + \beta \bar{u}^{(i)}_{t} - \left(\left(1-\beta\right) \bar{u}_{\mathrm{run}, t-1} + \beta \bar{u}_{t}\right)\right)^2 \\
    &= \frac{1}{n} \sum_{i=1}^n \left(\hat{\sigma}_{\mathrm{run}, t}^{(i)}\right)^2 + \frac{Kn}{Kn-1}\times\frac{1}{n\beta} \sum_{i=1}^n \left(\bar{u}_{\mathrm{run}, t}^{(i)} - \left(\frac{1}{n} \sum_{i=1}^n \bar{u}_{\mathrm{run}, t}^{(i)}\right)\right)^2 \\
    &= \frac{1}{n} \sum_{i=1}^n \left(\hat{\sigma}_{\mathrm{run}, t}^{(i)}\right)^2 + \frac{Kn}{Kn-1}\times\frac{1}{n\beta} \sum_{i=1}^n \left(\bar{u}_{\mathrm{run}, t}^{(i)} - \bar{u}_{\mathrm{run}, t}\right)^2 \tag{RV}\label{run_var}
\end{align*}

In \eqref{run_mean} and \eqref{run_var}, we show that the running mean and running variance computed in centralized setting can be express with the running mean and running variance of each clients. We use them in Algorithm~\ref{FBN}.

\section{Distributed Stochastic Gradient Descent With FBN}
\label{app:algoDSGD}

\begin{algorithm}[H]
 \caption{Distributed Stochastic Gradient Descent (DSGD) with model using FBN}
    \label{dsgd_fbn}
 \begin{algorithmic}[1]
     \STATE {\bfseries Initialization:} Server chooses initial model $\theta_1 \in \mathbb{R}^d$, running means $\bar{x}_{\mathrm{run}} = \mathbf{0}$, running variance $\hat{\sigma}_{\mathrm{run}}^2 = \mathbf{1}$, a {learning rate} $\eta \in \mathbb{R}^+$, batch size $b \geq 2$.
     \vspace{5pt}
     
     \FOR{$t = 1$ to $T$}
     \STATE Server broadcasts $\theta_t$ to all clients.\vspace{3pt}
     \FOR{\textbf{\textup{each}} client $i$ \textbf{in parallel}}
     \STATE Sample a mini-batch of data points $\left \{ \left(x_1^{(i)}, y_1^{(i)}\right ),  ..., \left(x_b^{(i)},y_b^{(i)} \right) \right \}$ of size $b$ from $S_i$
     \vspace{5pt}
     \STATE Compute the forward pass $\left\{ F_{\theta_t}\left( x_1^{(i)}\right),  ..., F_{\theta_t}\left(x_b^{(i)} \right )\right\}$ using FBN for each layer's input 
     \vspace{5pt}
     \STATE Compute gradient $g_t^{(i)} = \frac{1}{b} \sum_{j=1}^b\nabla_{\theta_t} \ell\left(F_{\theta_t}\left(x_j^{(i)} \right),y_j^{(i)} \right)$ 
     \vspace{3pt}
     \STATE Send $g_t^{(i)}$ to the Server
     \vspace{3pt}
     \ENDFOR
     \STATE Server aggregates the gradients $g_t = \frac{1}{n} \sum_{i=1}^n g_t^{(i)}$
     \vspace{3pt}
     \STATE Server computes the updated model $\theta_{t+1} = \theta_{t} - \eta \, g_t$
     \ENDFOR
     \vspace{5pt}
     \STATE {\bfseries Output:}  Server outputs $\theta_{T}$
 \end{algorithmic}
 \end{algorithm}

\section{Additional Experimental results}

In this section we present the additional information on the setup of our experiments presented in Section~\ref{sec_fed_setting} and \ref{sec:byz}and show additional results.

\subsection{Experimental setup}
\label{exp_setup}

The model used in our experiments is presented in Table~\ref{tab:cnn_cifar}.

All the hyperparameters are listed in Table~\ref{tab:exp3}

\begin{table}[h]
\small
    \centering
    \setlength{\tabcolsep}{10pt} 
    \renewcommand{\arraystretch}{2} 
    \begin{tabular}{|c|c|}
        \hline
        First Layer & Convolution : in = 3, out = 64, kernel\_size = 3, padding = 1, stride = 1\\
         & ReLU \\
         & BatchtNorm2d \\
         \hline
         Second Layer & Convolution : in = 64, out = 64, kernel\_size = 3, padding=1, stride = 1\\
         & ReLU \\
         & BatchtNorm2d \\
         & MaxPool(2,2) \\
         & DropOut(0.25) \\
         \hline
         Third Layer & Convolution : in = 64, out = 128, kernel\_size = 3, padding=1, stride = 1\\
         & ReLU \\
         & BatchtNorm2d \\
         \hline
         Fourth Layer & Convolution : in = 128, out = 128, kernel\_size = 3, padding=1, stride = 1\\
         & ReLU \\
         & BatchtNorm2d \\
         & MaxPool(2,2) \\
         & DropOut(0.25) \\
         \hline
         Fifth Layer & Linear : in = $8192$, out = $128$\\
         & ReLU \\
        \hline
         Sixth Layer & Linear : in = $128$, out = $10$\\
         & Log Softmax \\
        \hline
    \end{tabular}
    \vspace{3pt}
    \caption{CNN Architecture for Cifar10}
    \label{tab:cnn_cifar}
\end{table}

\begin{table}[ht]
\small
    \centering
    \setlength{\tabcolsep}{10pt} 
    \renewcommand{\arraystretch}{2} 
    \begin{tabular}{|c|c|}
        \hline
        Model &  CNN (architecture presented in Table~\ref{tab:cnn_cifar})\\
        \hline
        Algorithm & DSGD in Section~\ref{sec_fed_setting} and Robust-DSGD in Section~\ref{sec:byz}\\
        \hline
        Number of steps & $T=3000$ \\
        \hline
        Learning rate & 
            $\begin{matrix}
                        &0.1 & \text{if} & 0 &\leq & T & <& 1000\\
                        &0.05 & \text{if} & 1000 &\leq & T & <& 2000\\
                        &0.033 & \text{if} & 2000 &\leq & T & <& 3000\\
                        \end{matrix}$ \\
        \hline
        Batch size & 50 \\
        \hline
        Client momentum & 0.99 \\
        \hline
        Loss function & Negative Log Likelihood (NLL)\\ 
        \hline
    \end{tabular}
    \vspace{3pt}
    \caption{Experimental setup}
    \label{tab:exp3}
\end{table}

We list below the hardware components used for our experiments:
\begin{itemize}
    \item 1 Intel(R) Core(TM) i7-8700K CPU @ 3.70GHz
    \item 2 Nvidia GeForce GTX 1080 Ti
    \item 64 GB of RAM
\end{itemize}

\subsection{Additional experiments on \fbn{} using Dirichlet Heterogeneity metric}
In this section, we present the same results presented in the paper but with a different heterogeneity model. More precisely, we use the Dirichlet method to distribute the data across the clients, as presented in \cite{dirichlet}. This method is parameterized with a real number $\alpha \in \mathbb{R}_+^*$. The closer $\alpha$ is to zero, the more heterogeneous the data.

\begin{figure}[ht!]
    \centering
    \subfloat[Final acc. vs. Heterogeneity]{\includegraphics[width=0.33\textwidth]{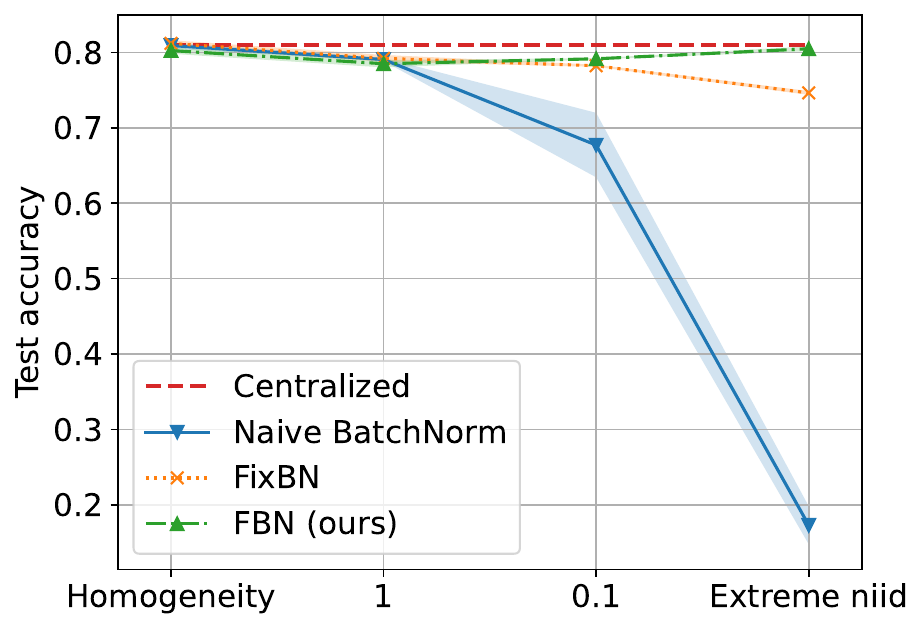}}
    \label{fig_comparison_centralized:final_2}
    \subfloat[$\gamma=0.01$]{\includegraphics[width=0.32\textwidth]{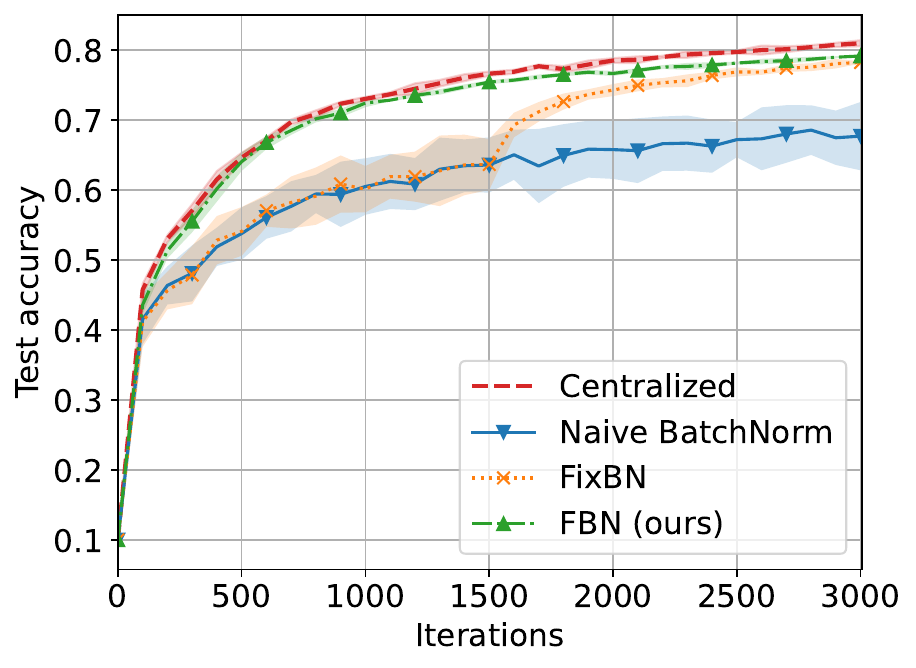}}
    \label{fig_comparison_centralized:0_01_2}
    \subfloat[$\gamma = 0$]{\includegraphics[width=0.32\textwidth]{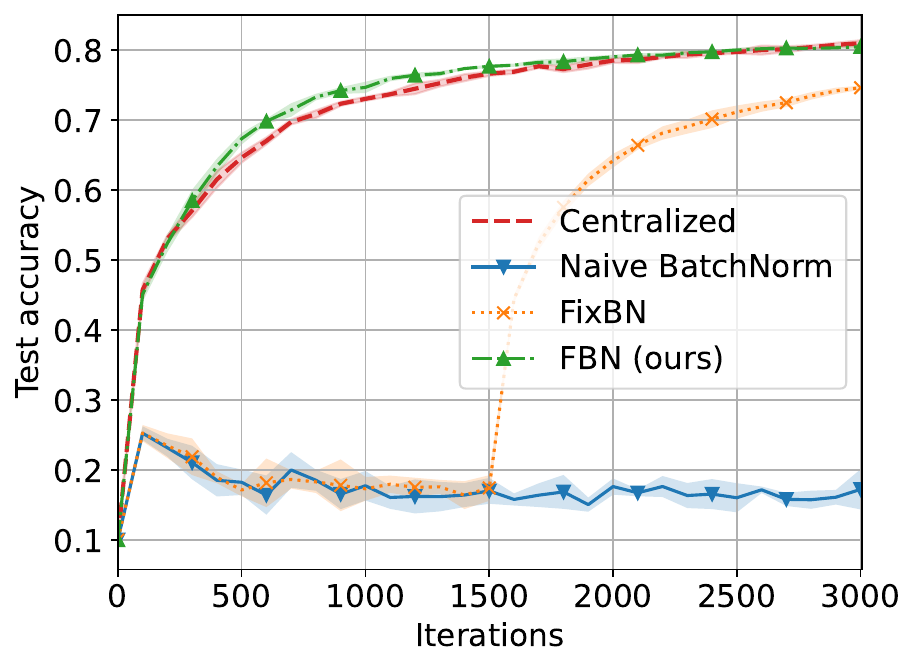}
    \label{fig_comparison_centralized:0_2}}
    \caption{Final accuracy of DSGD using $\{$\fbn{}, FixBN, Naive \bn{}$\}$ vs heterogeneity (a), evolution of test accuracy for $\alpha = 0.1$ (b) and in extreme heterogeneity ($\gamma=0$) (c).}
    \label{fig_comparison_centralized_2}
\end{figure}

\clearpage
\subsection{Comprehensive Results on Robustness of \fbn{}}\label{app_byz_setting}

\subsubsection{Results with $\gamma$-similarity}

\begin{figure*}[ht!]
    \centering
    \subfloat[$\gamma = 0.5$]{\includegraphics[width=0.49\textwidth]{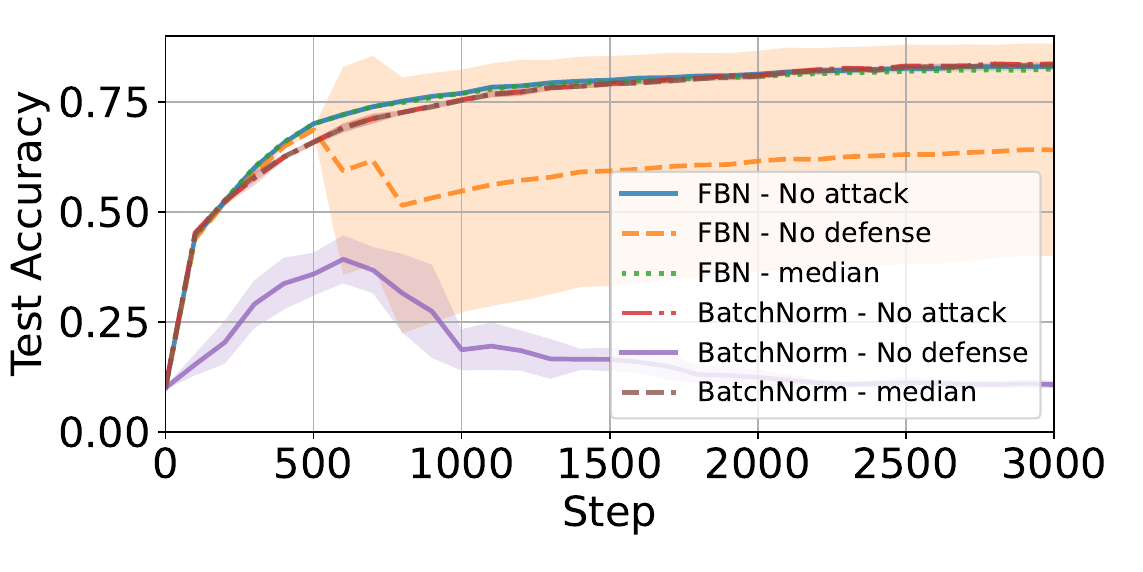}}
    \subfloat[$\gamma = 0.1$]{\includegraphics[width=0.49\textwidth]{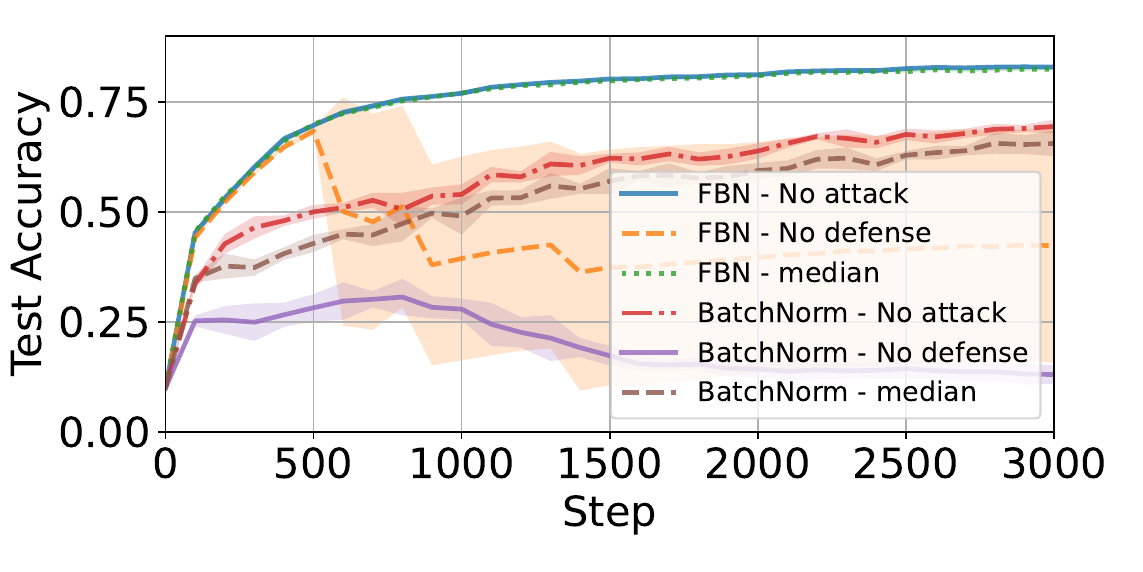}}\\
     \subfloat[$\gamma = 0.01$]{\includegraphics[width=0.49\textwidth]{figures/cifar10_SF_f=3_model=cnn_cifar_old_lr=0.1_heterogeneity=gamma_0.01_median.pdf}}
    \subfloat[extreme heterogeneity]{\includegraphics[width=0.49\textwidth]{figures/cifar10_SF_f=3_model=cnn_cifar_old_lr=0.1_heterogeneity=extreme_median.pdf}}
    \caption{Performance of \fbn{} vs. \bn{} on CIFAR-10 in four heterogeneity settings, using the coordinate-wise median~\cite{yin2018byzantine} aggregation.
    Among a total of 10 clients, we consider $f = 3$ adversarial clients executing \textbf{SF}~\cite{allen2020byzantine}.}
\label{fig_byz_app_SF_median}
\end{figure*}

\begin{figure*}[ht!]
    \centering
    \subfloat[$\gamma = 0.5$]{\includegraphics[width=0.49\textwidth]{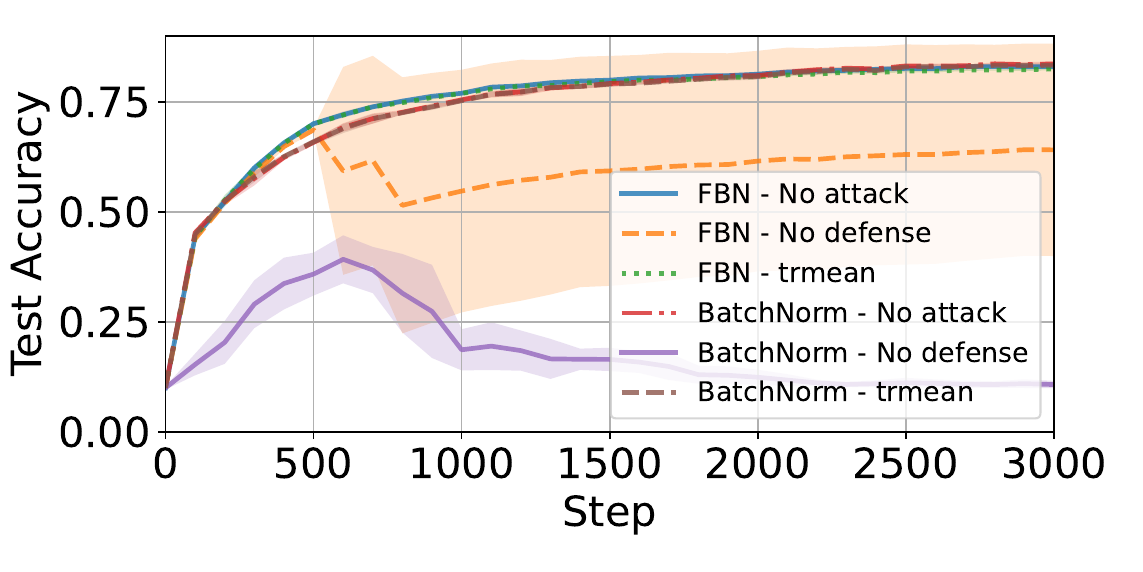}}
    \subfloat[$\gamma = 0.1$]{\includegraphics[width=0.49\textwidth]{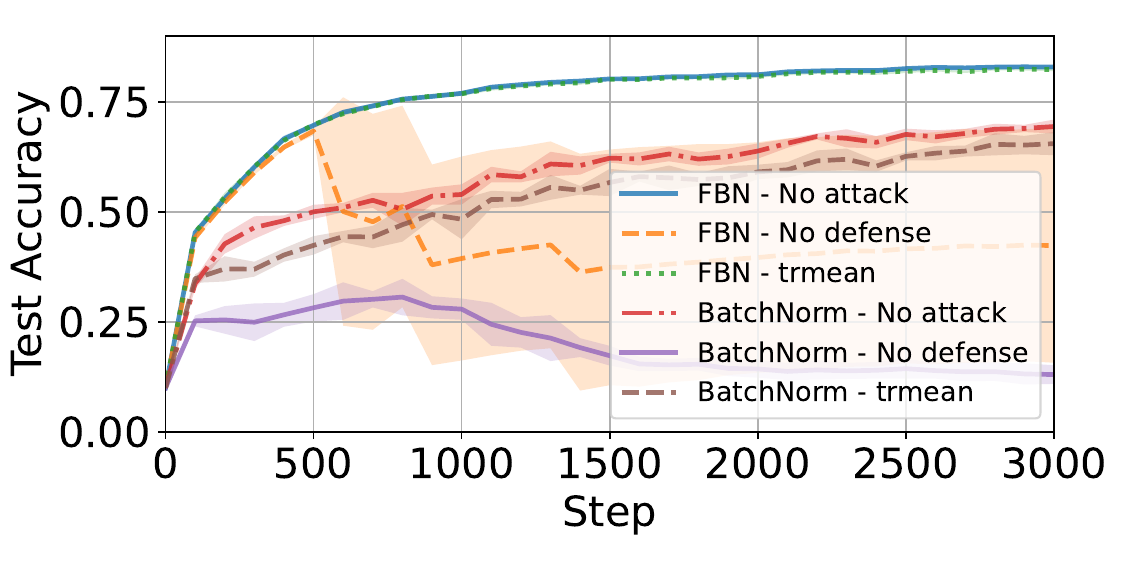}}\\
     \subfloat[$\gamma = 0.01$]{\includegraphics[width=0.49\textwidth]{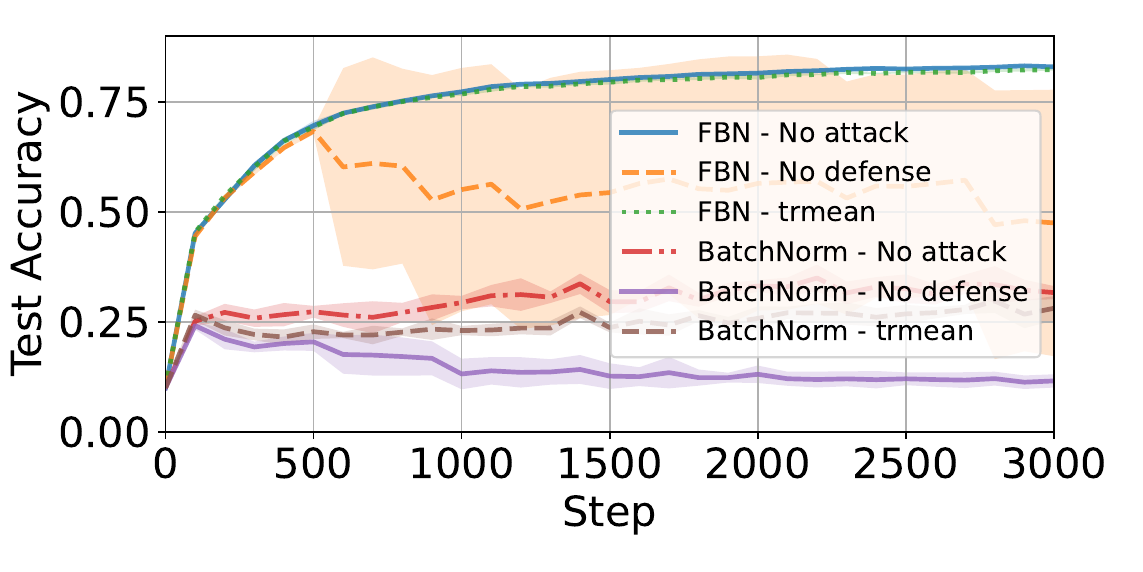}}
    \subfloat[extreme heterogeneity]{\includegraphics[width=0.49\textwidth]{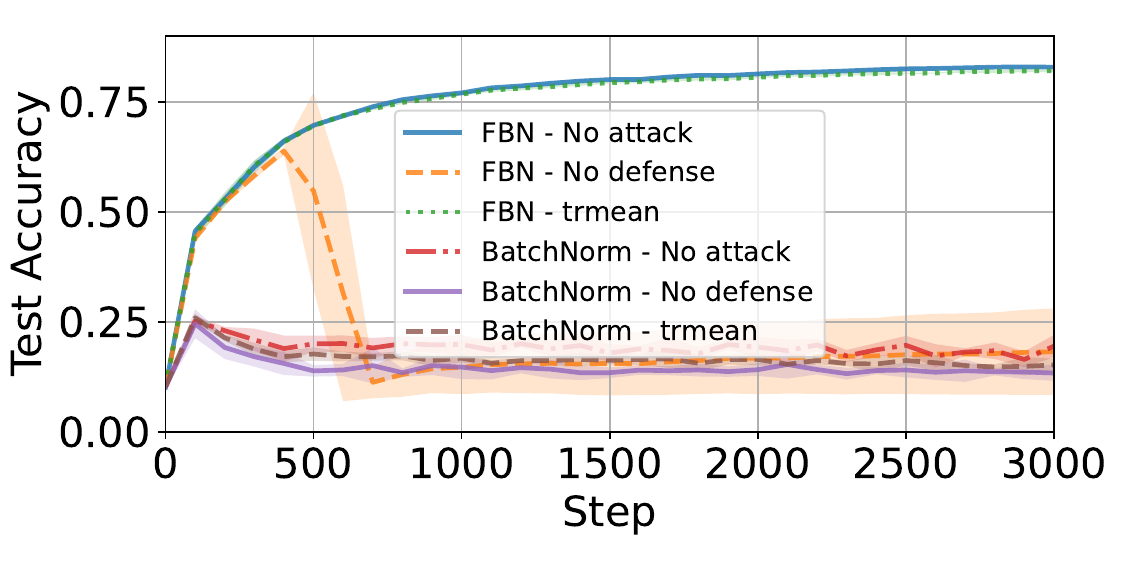}}
    \caption{Performance of \fbn{} vs. \bn{} on CIFAR-10 in four heterogeneity settings, using the coordinate-wise trimmed mean~\cite{yin2018byzantine} aggregation.
    Among a total of 10 clients, we consider $f = 3$ adversarial clients executing \textbf{SF}~\cite{allen2020byzantine}.}
\label{fig_byz_app_SF_trmean}
\end{figure*}

\begin{figure*}[ht!]
    \centering
    \subfloat[$\gamma = 0.5$]{\includegraphics[width=0.49\textwidth]{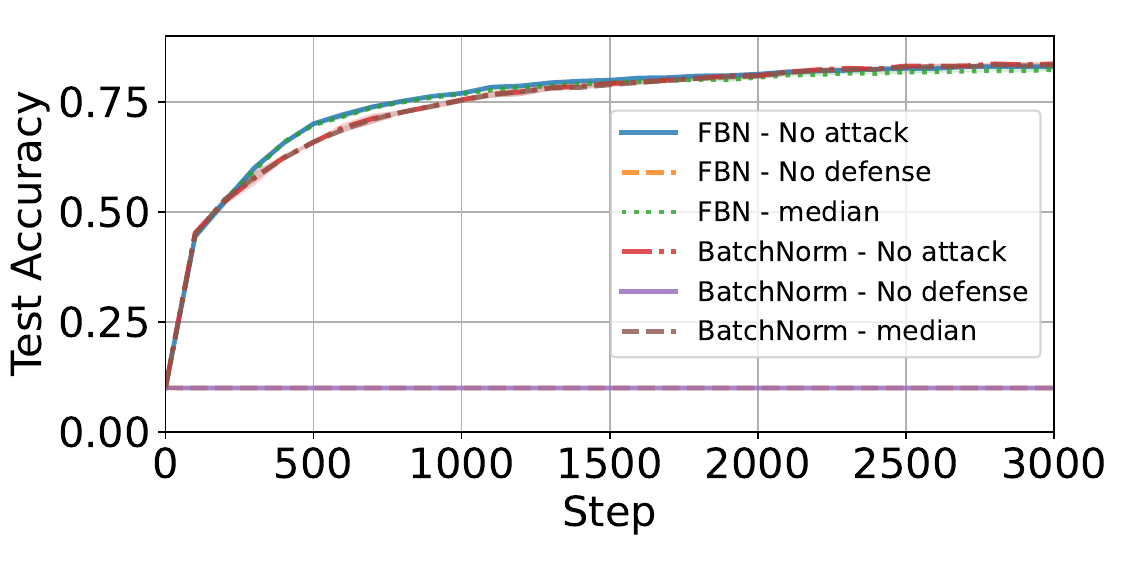}}
    \subfloat[$\gamma = 0.1$]{\includegraphics[width=0.49\textwidth]{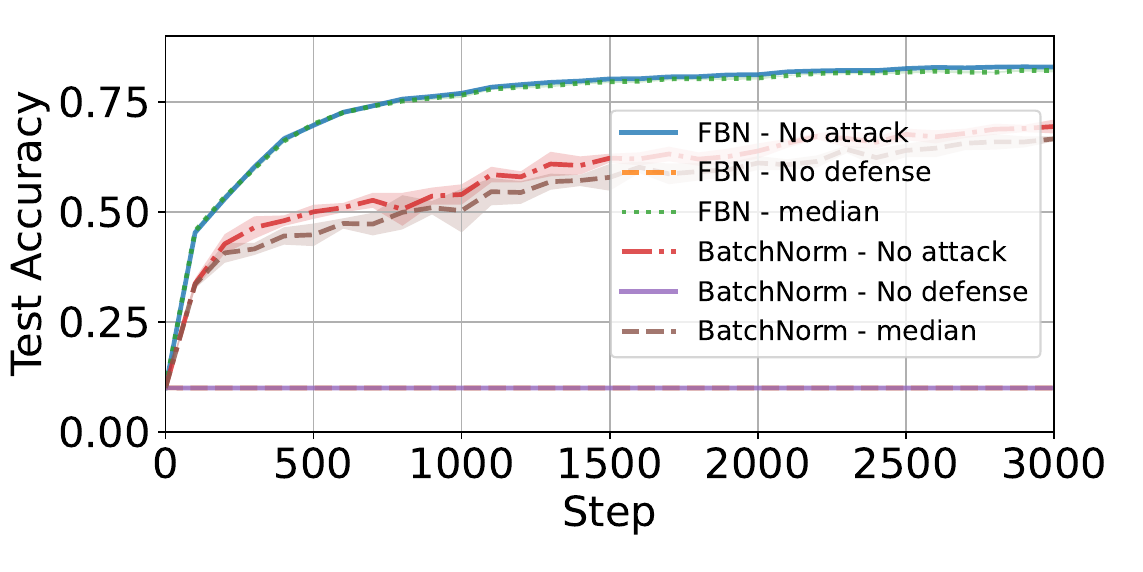}}\\
     \subfloat[$\gamma = 0.01$]{\includegraphics[width=0.49\textwidth]{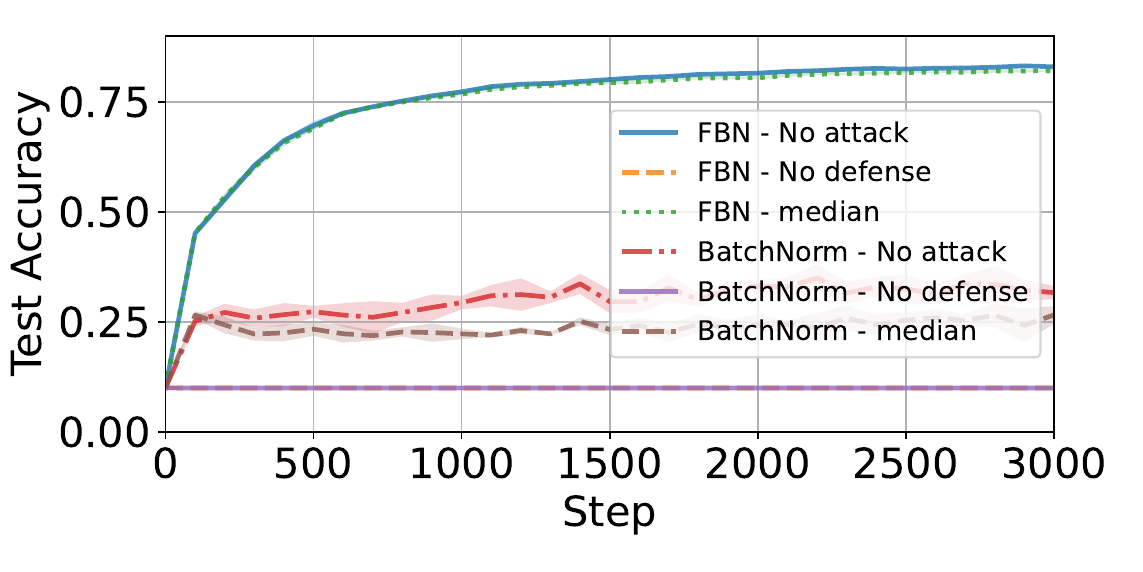}}
    \subfloat[extreme heterogeneity]{\includegraphics[width=0.49\textwidth]{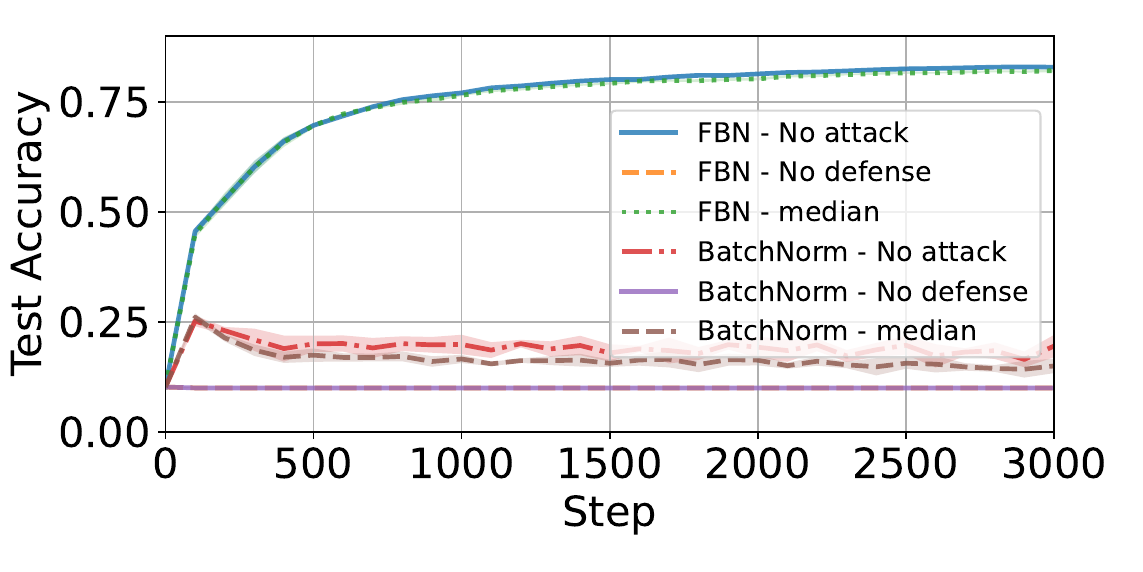}}
    \caption{Performance of \fbn{} vs. \bn{} on CIFAR-10 in four heterogeneity settings, using the coordinate-wise median~\cite{yin2018byzantine} aggregation.
    Among a total of 10 clients, we consider $f = 3$ adversarial clients executing \textbf{FOE}~\cite{xie2020fall}.}
\label{fig_byz_app_FOE_median}
\end{figure*}

\begin{figure*}[ht!]
    \centering
    \subfloat[$\gamma = 0.5$]{\includegraphics[width=0.49\textwidth]{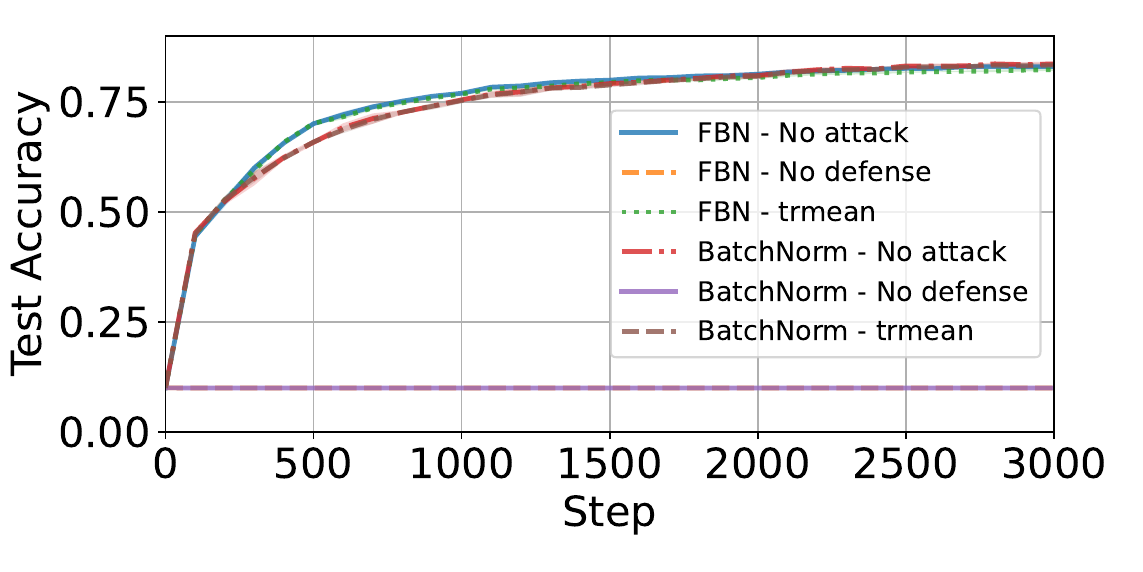}}
    \subfloat[$\gamma = 0.1$]{\includegraphics[width=0.49\textwidth]{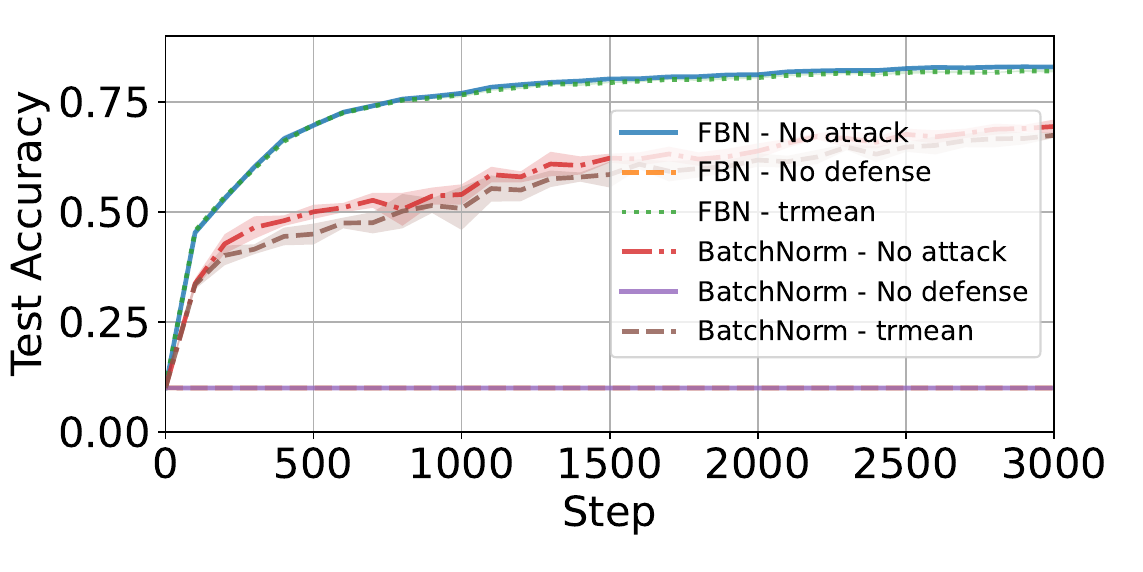}}\\
     \subfloat[$\gamma = 0.01$]{\includegraphics[width=0.49\textwidth]{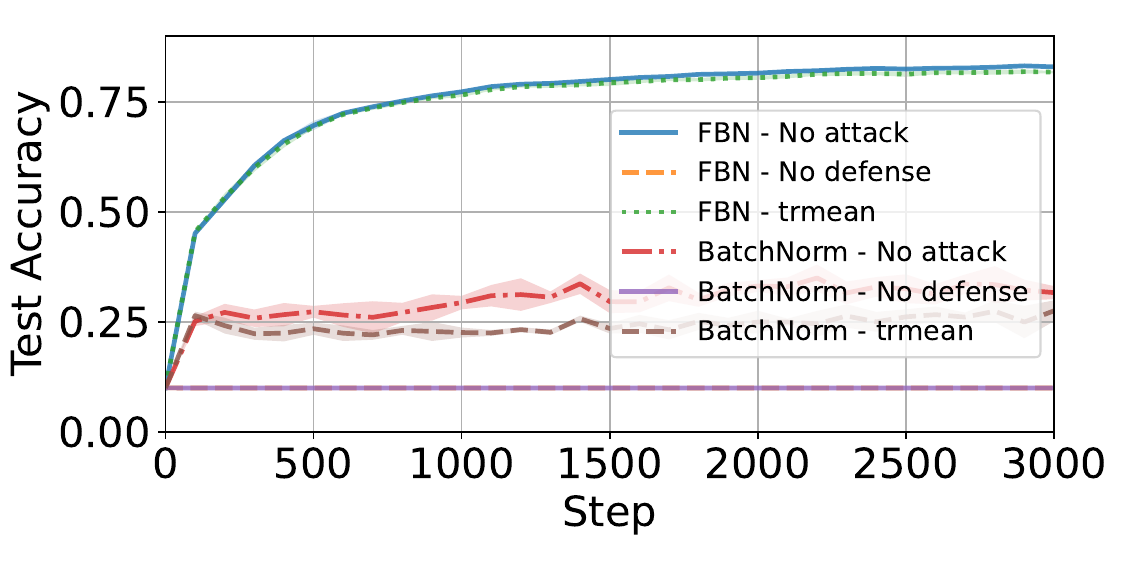}}
    \subfloat[extreme heterogeneity]{\includegraphics[width=0.49\textwidth]{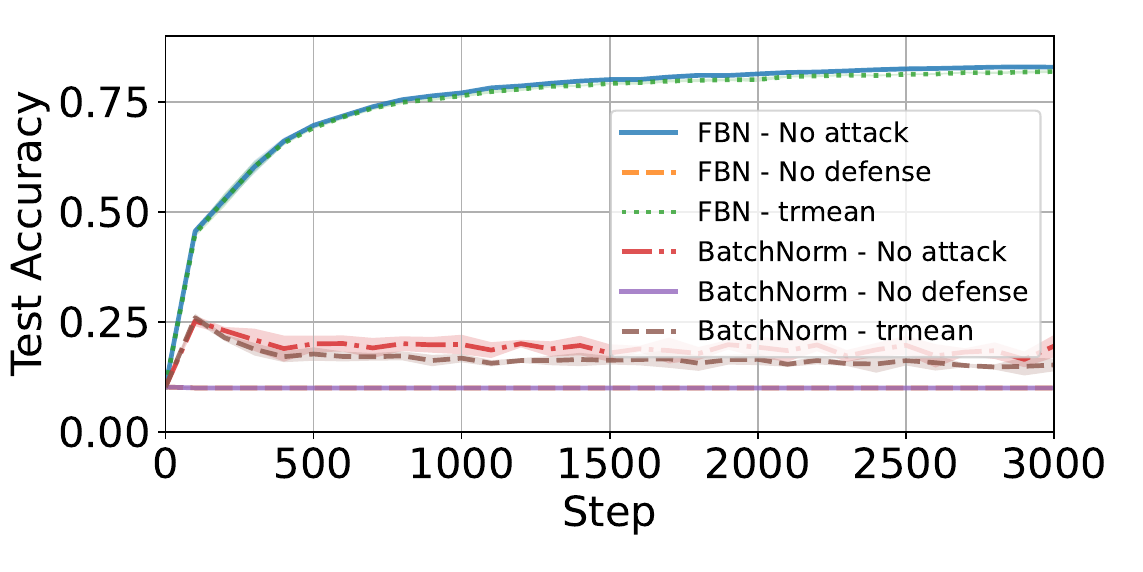}}
    \caption{Performance of \fbn{} vs. \bn{} on CIFAR-10 in four heterogeneity settings, using the coordinate-wise trimmed mean~\cite{yin2018byzantine} aggregation.
    Among a total of 10 clients, we consider $f = 3$ adversarial clients executing \textbf{FOE}~\cite{xie2020fall}.}
\label{fig_byz_app_FOE_trmean}
\end{figure*}

\begin{figure*}[ht!]
    \centering
    \subfloat[$\gamma = 0.5$]{\includegraphics[width=0.49\textwidth]{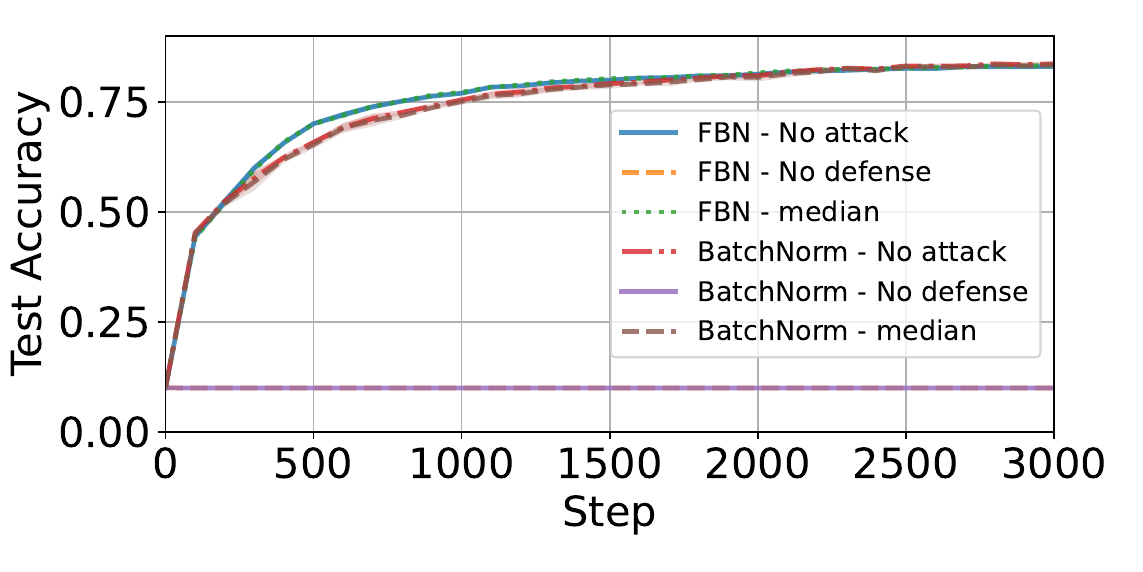}}
    \subfloat[$\gamma = 0.1$]{\includegraphics[width=0.49\textwidth]{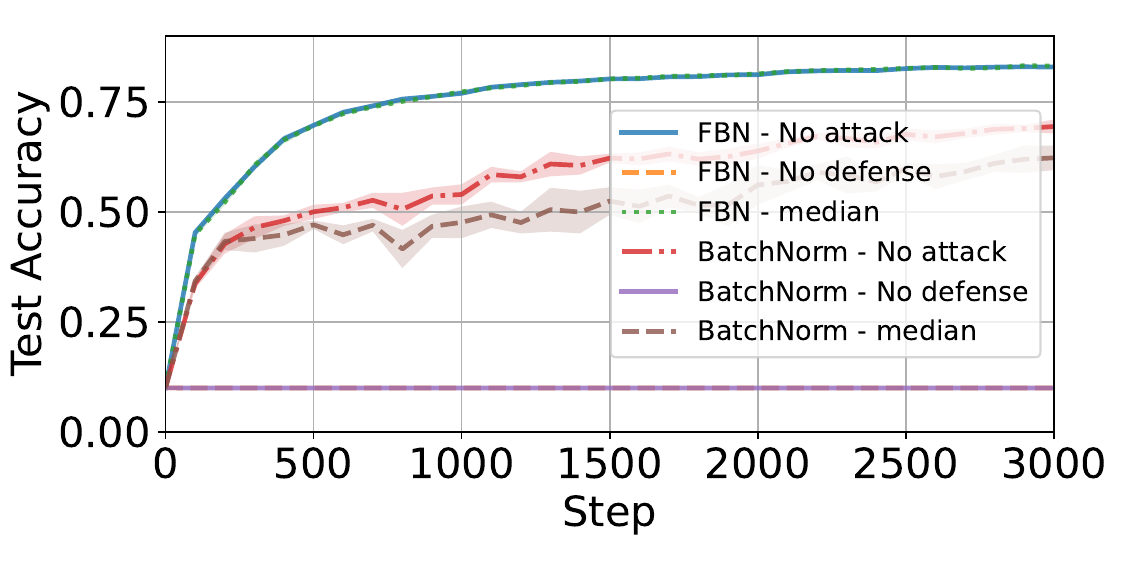}}\\
     \subfloat[$\gamma = 0.01$]{\includegraphics[width=0.49\textwidth]{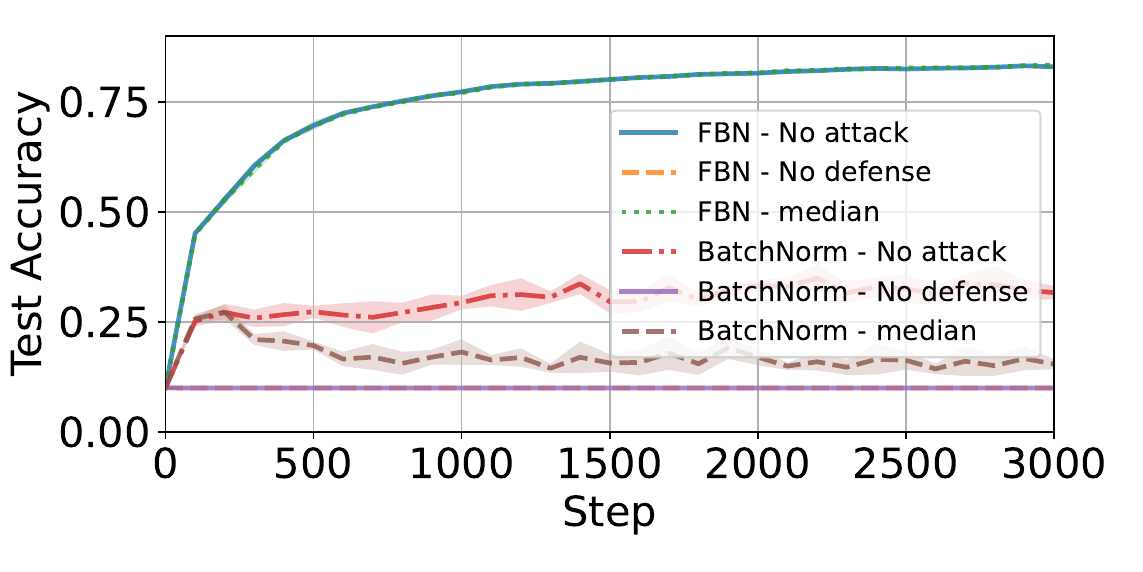}}
    \subfloat[extreme heterogeneity]{\includegraphics[width=0.49\textwidth]{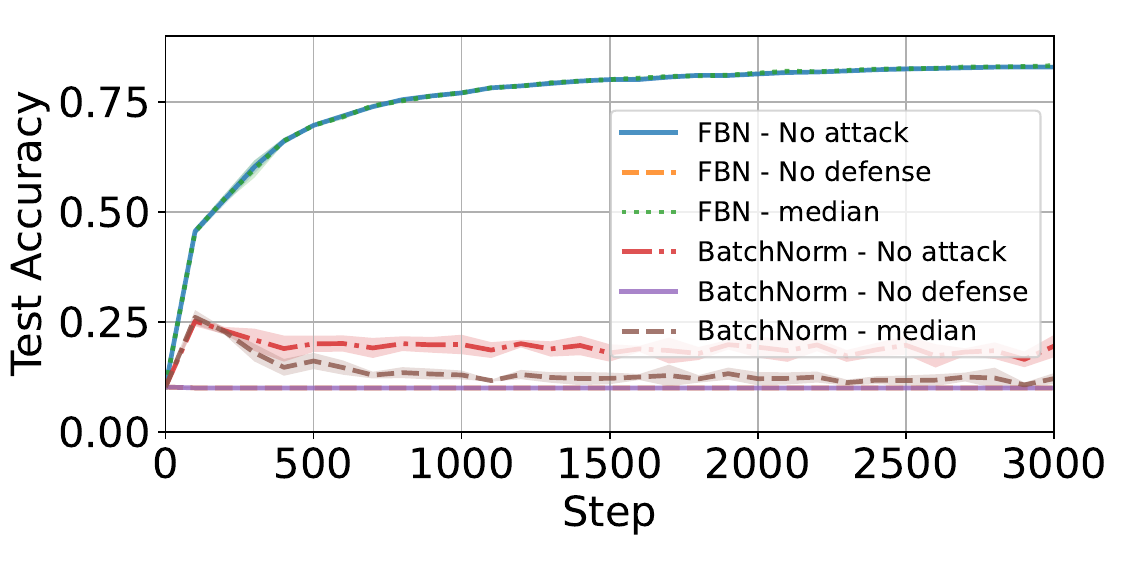}}
    \caption{Performance of \fbn{} vs. \bn{} on CIFAR-10 in four heterogeneity settings, using the coordinate-wise median~\cite{yin2018byzantine} aggregation.
    Among a total of 10 clients, we consider $f = 3$ adversarial clients executing \textbf{ALIE}~\cite{baruch2019alittle}.}
\label{fig_byz_app_ALIE_median}
\end{figure*}

\begin{figure*}[ht!]
    \centering
    \subfloat[$\gamma = 0.5$]{\includegraphics[width=0.49\textwidth]{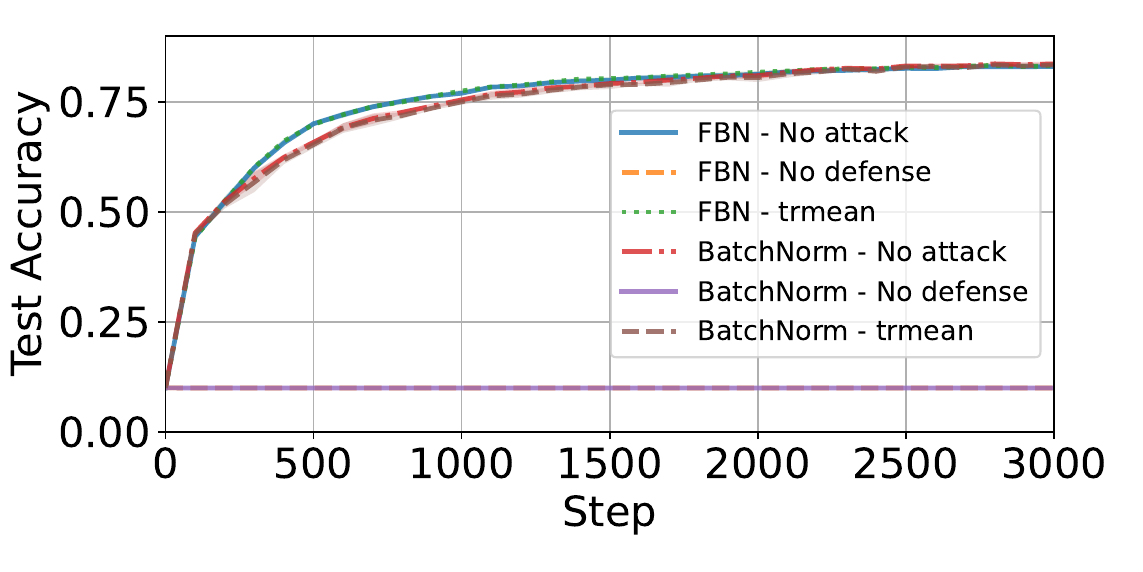}}
    \subfloat[$\gamma = 0.1$]{\includegraphics[width=0.49\textwidth]{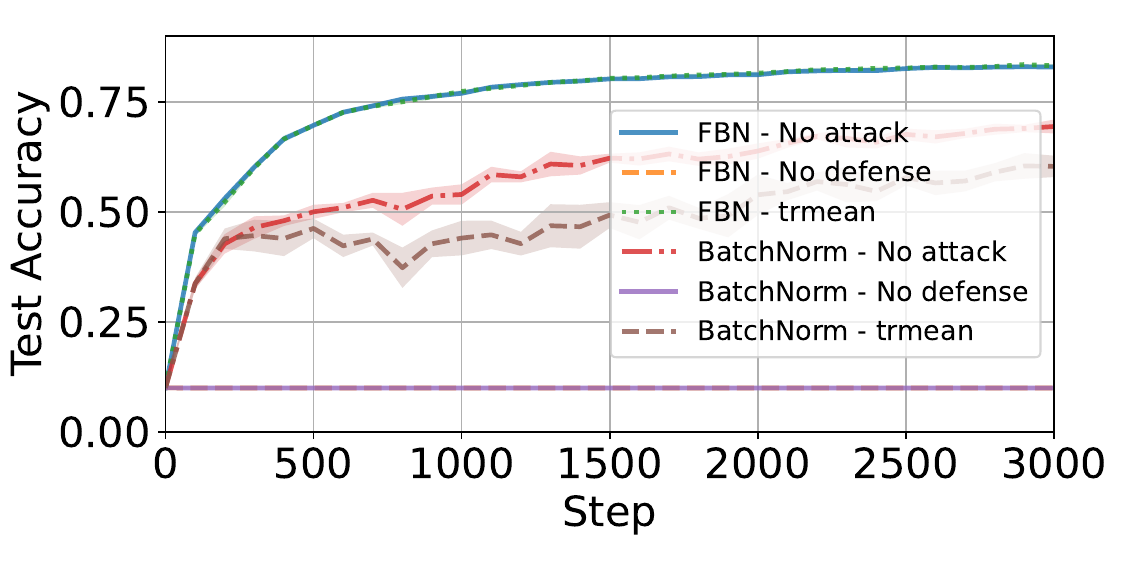}}\\
     \subfloat[$\gamma = 0.01$]{\includegraphics[width=0.49\textwidth]{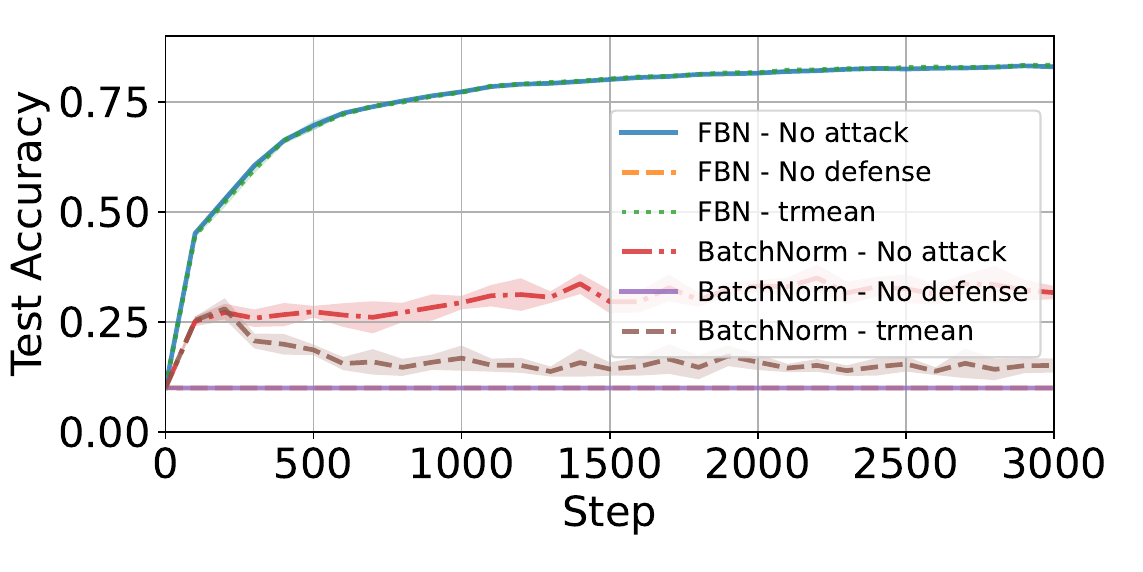}}
    \subfloat[extreme heterogeneity]{\includegraphics[width=0.49\textwidth]{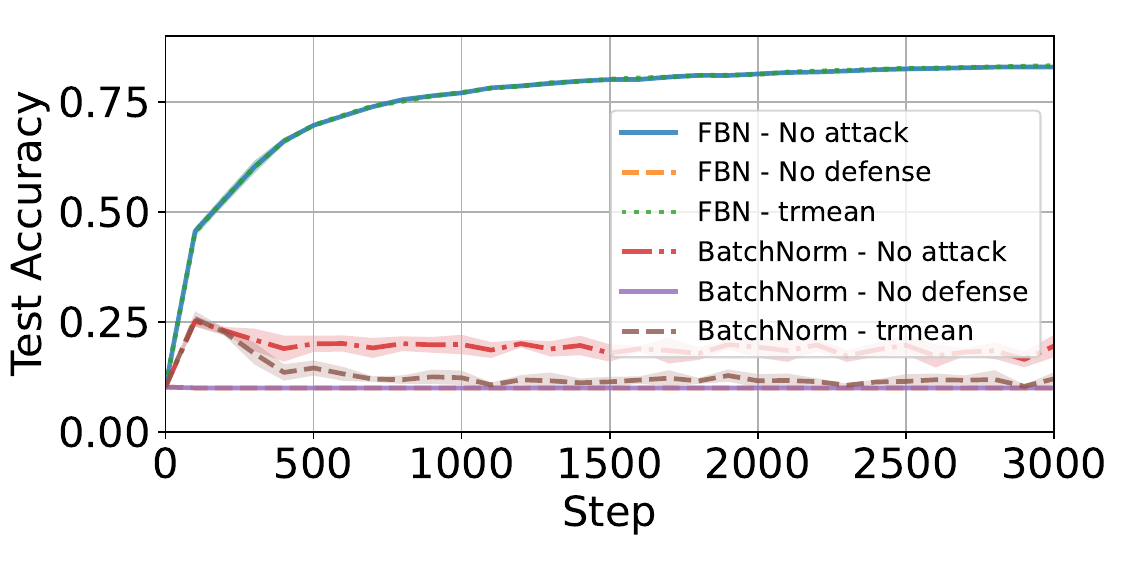}}
    \caption{Performance of \fbn{} vs. \bn{} on CIFAR-10 in four heterogeneity settings, using the coordinate-wise trimmed mean~\cite{yin2018byzantine} aggregation.
    Among a total of 10 clients, we consider $f = 3$ adversarial clients executing \textbf{ALIE}~\cite{baruch2019alittle}.}
\label{fig_byz_app_ALIE_trmean}
\end{figure*}

\clearpage
\subsubsection{Results with Dirichlet}

\begin{figure*}[ht!]
    \centering
    \subfloat[ALIE - $\alpha = 0.1$]{\includegraphics[width=0.49\textwidth]{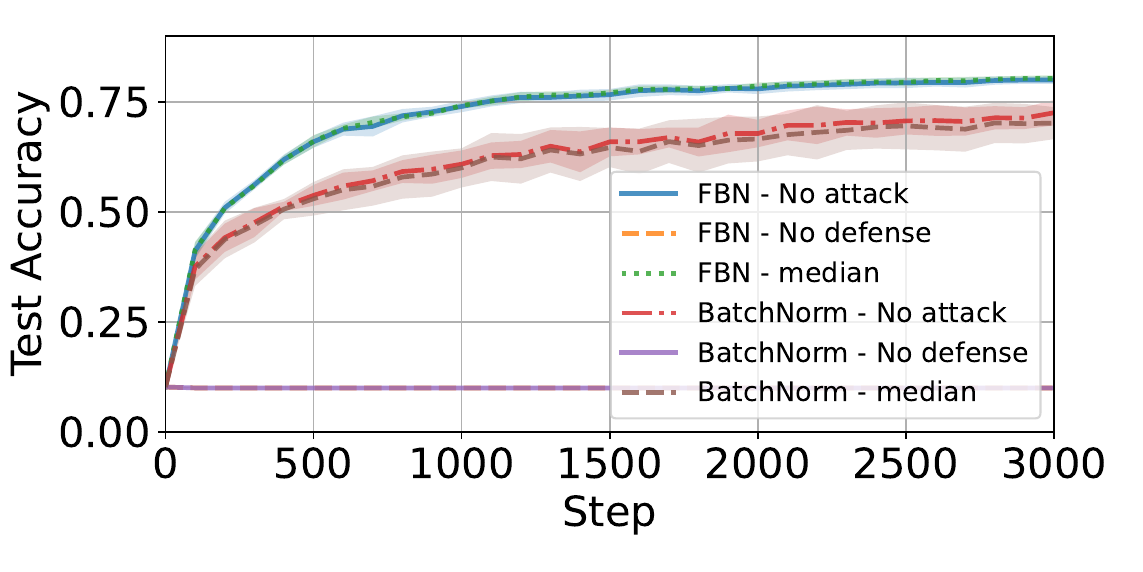}}
    \subfloat[ALIE - $\alpha = 1$]{\includegraphics[width=0.49\textwidth]{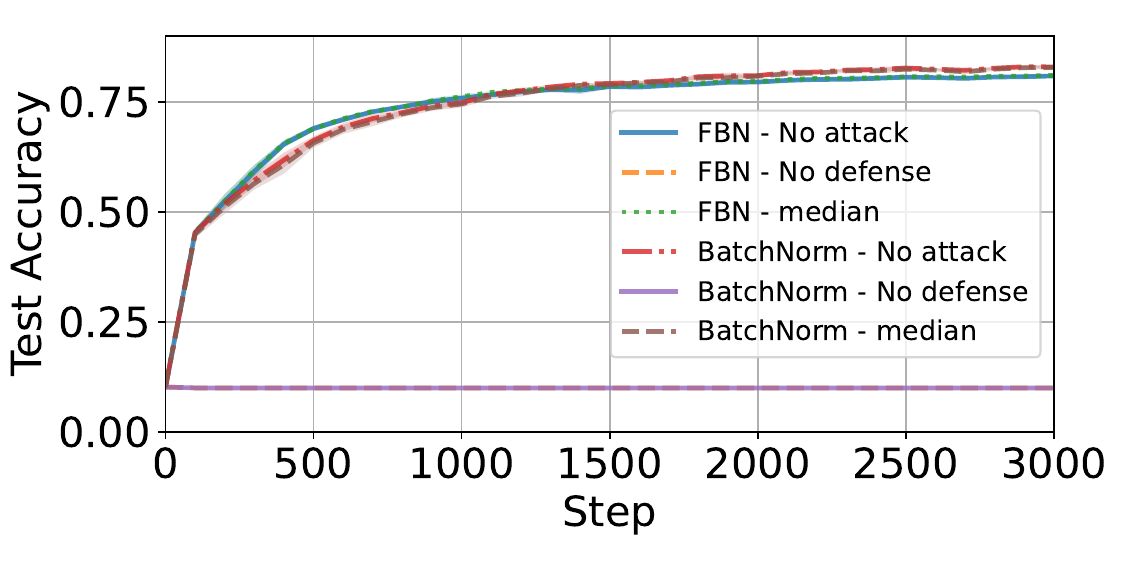}}\\
     \subfloat[FOE - $\alpha = 0.1$]{\includegraphics[width=0.49\textwidth]{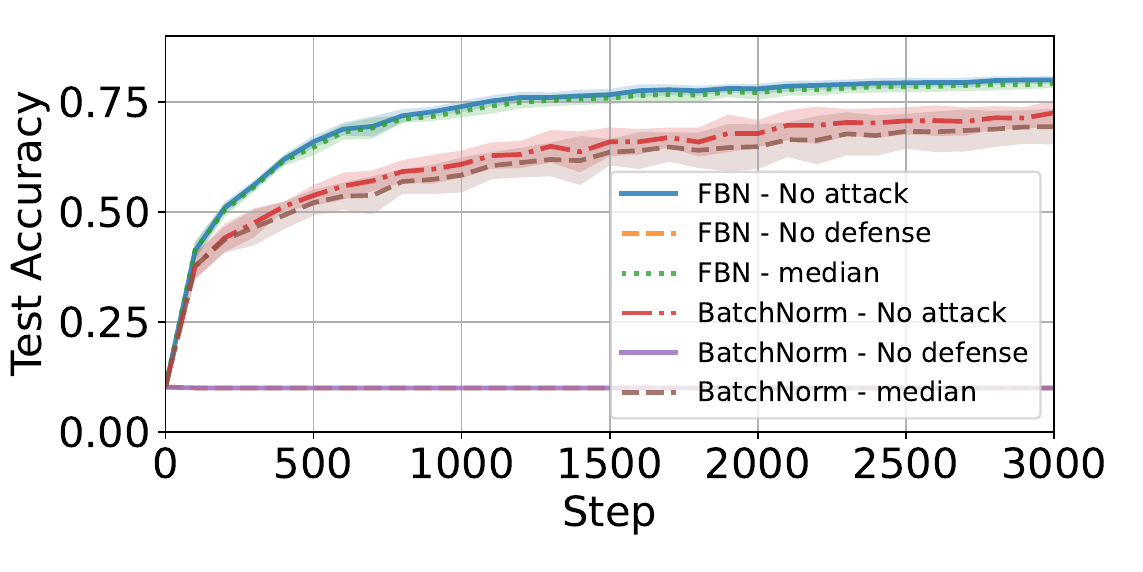}}
    \subfloat[FOE - $\alpha = 1$]{\includegraphics[width=0.49\textwidth]{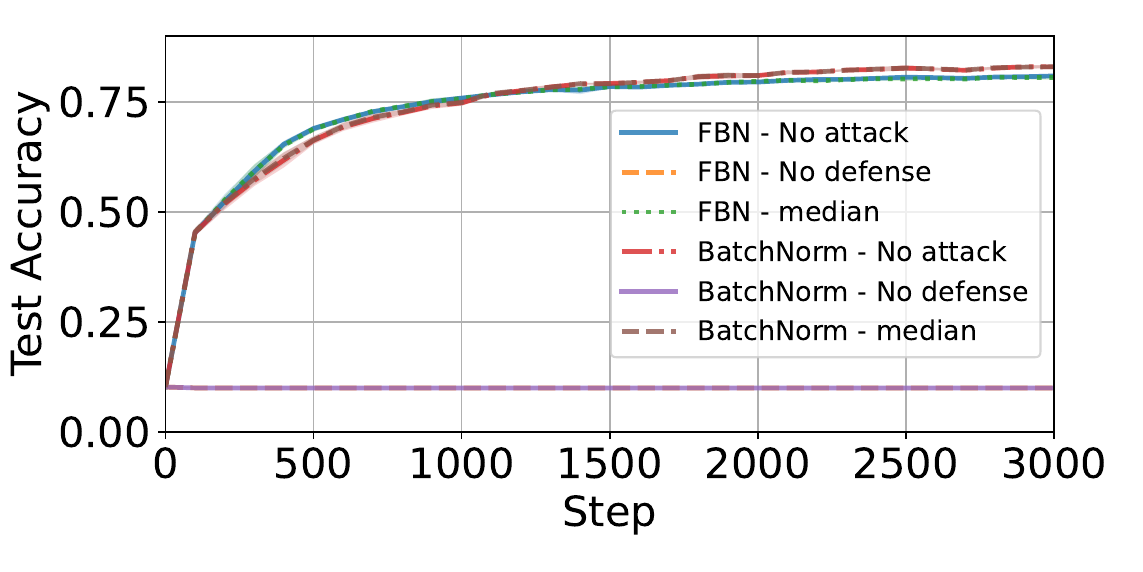}}\\
    \subfloat[SF - $\alpha = 0.1$]{\includegraphics[width=0.49\textwidth]{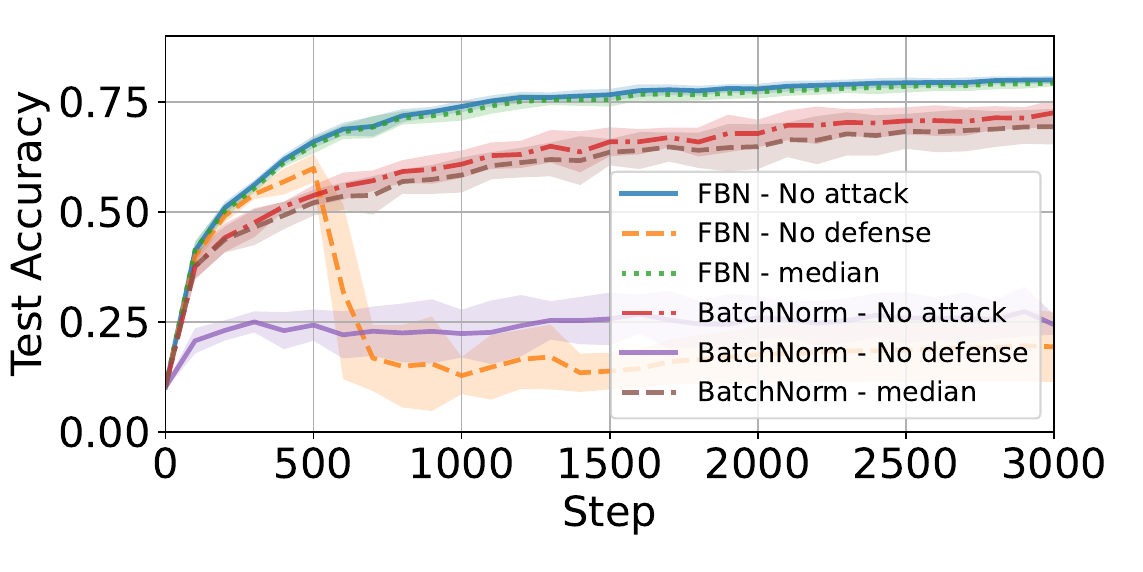}}
    \subfloat[SF - $\alpha = 1$]{\includegraphics[width=0.49\textwidth]{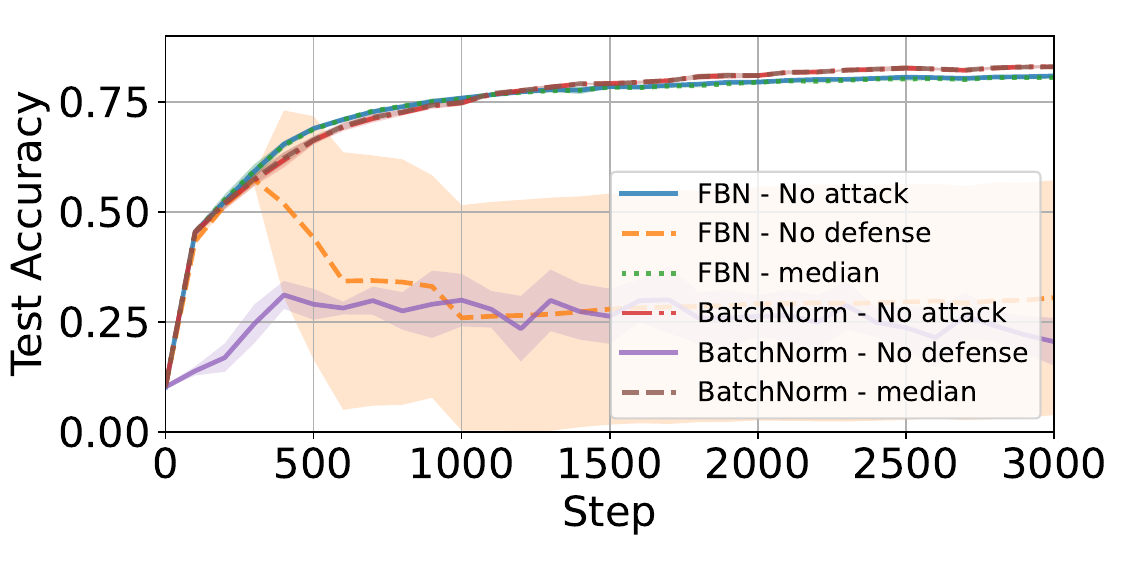}}
    \caption{Performance of \fbn{} vs. \bn{} on CIFAR-10 in two heterogeneity settings ($\alpha = 0.1$ and $\alpha = 1$), using the median~\cite{yin2018byzantine} aggregation.
    Among a total of 10 clients, we consider $f = 3$ adversarial clients executing the \textbf{ALIE}~\cite{baruch2019alittle}, \textbf{FOE}~\cite{xie2020fall}, and \textbf{SF}~\cite{allen2020byzantine} attacks.}
\label{fig_byz_app_dirichlet_median}
\end{figure*}

\begin{figure*}[ht!]
    \centering
    \subfloat[ALIE - $\alpha = 0.1$]{\includegraphics[width=0.49\textwidth]{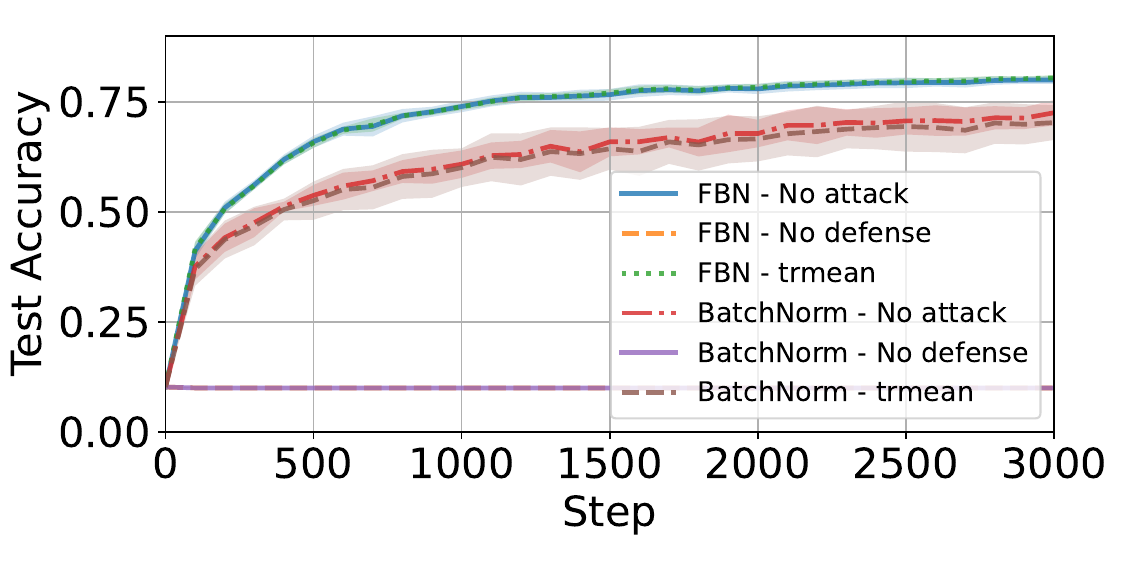}}
    \subfloat[ALIE - $\alpha = 1$]{\includegraphics[width=0.49\textwidth]{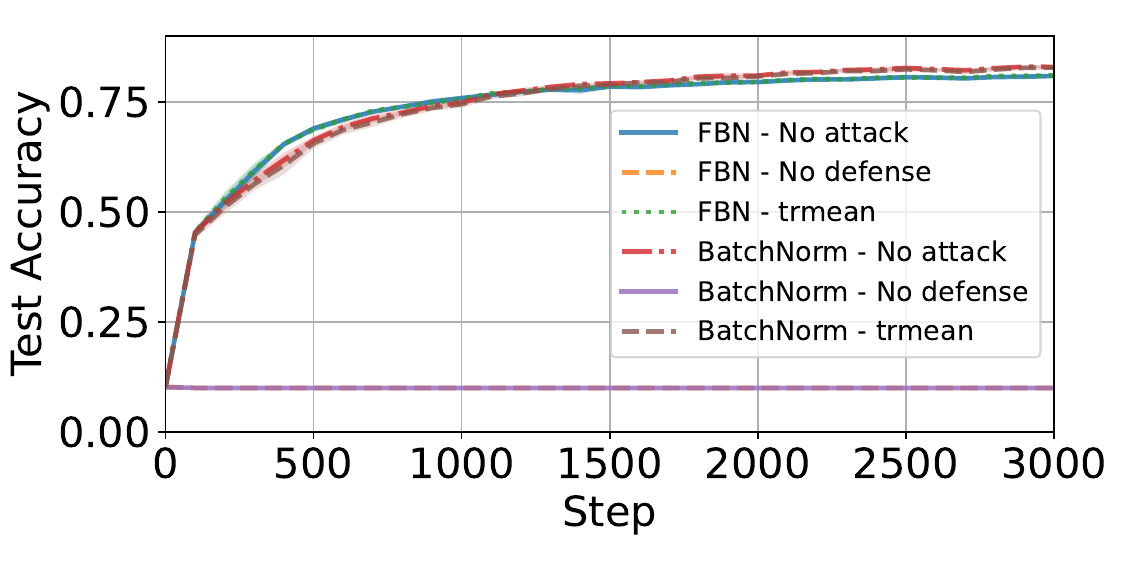}}\\
     \subfloat[FOE - $\alpha = 0.1$]{\includegraphics[width=0.49\textwidth]{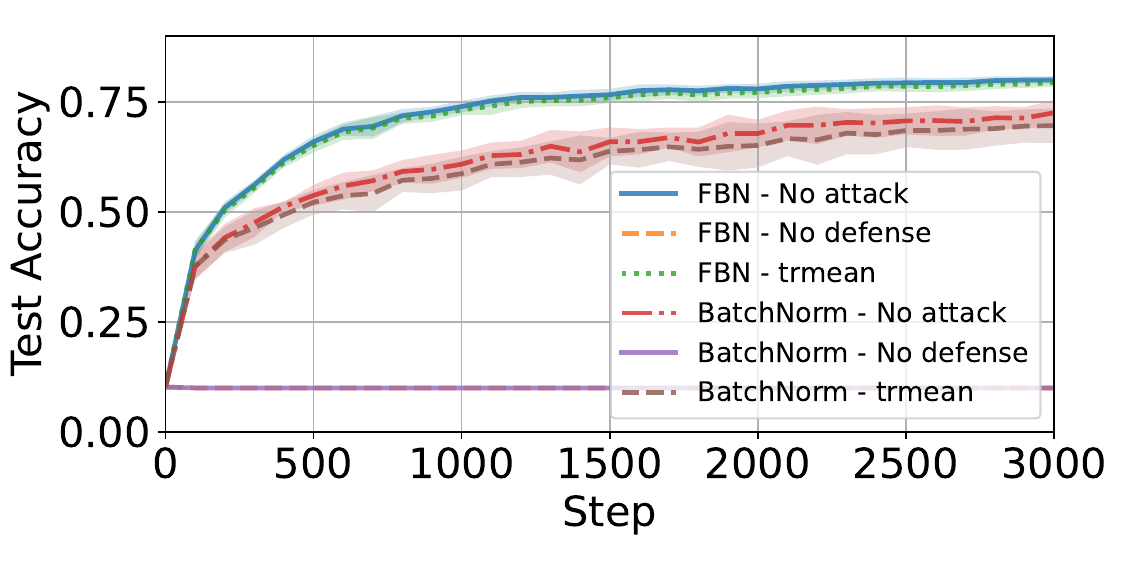}}
    \subfloat[FOE - $\alpha = 1$]{\includegraphics[width=0.49\textwidth]{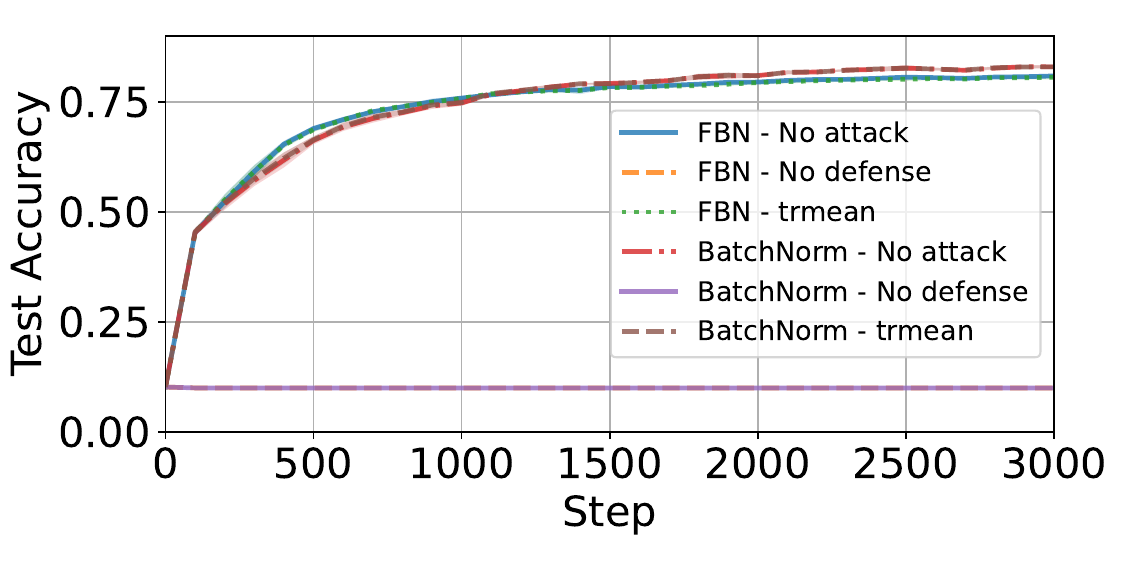}}\\
    \subfloat[SF - $\alpha = 0.1$]{\includegraphics[width=0.49\textwidth]{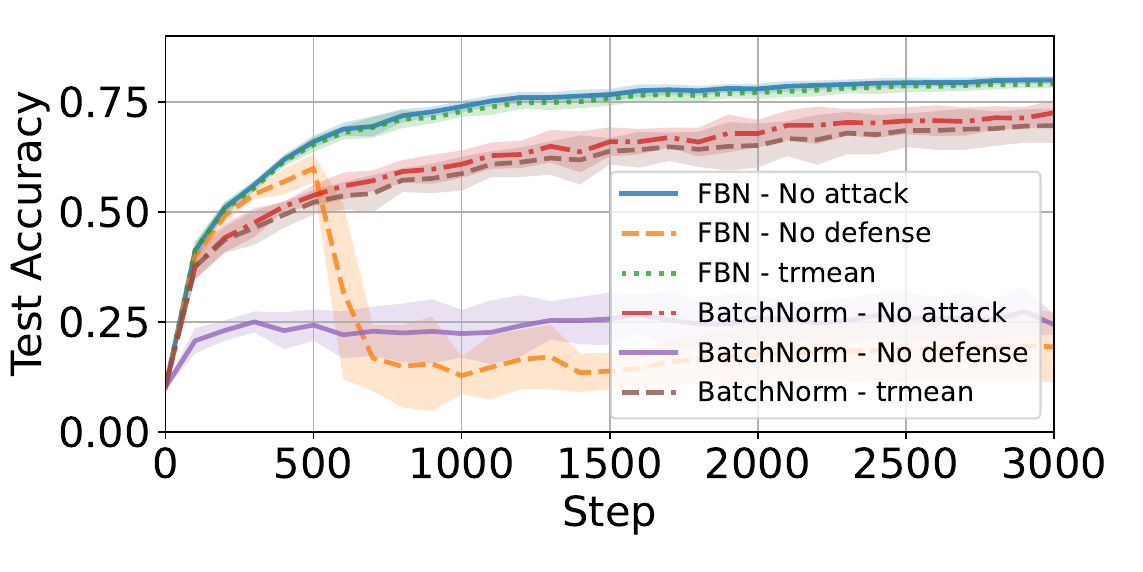}}
    \subfloat[SF - $\alpha = 1$]{\includegraphics[width=0.49\textwidth]{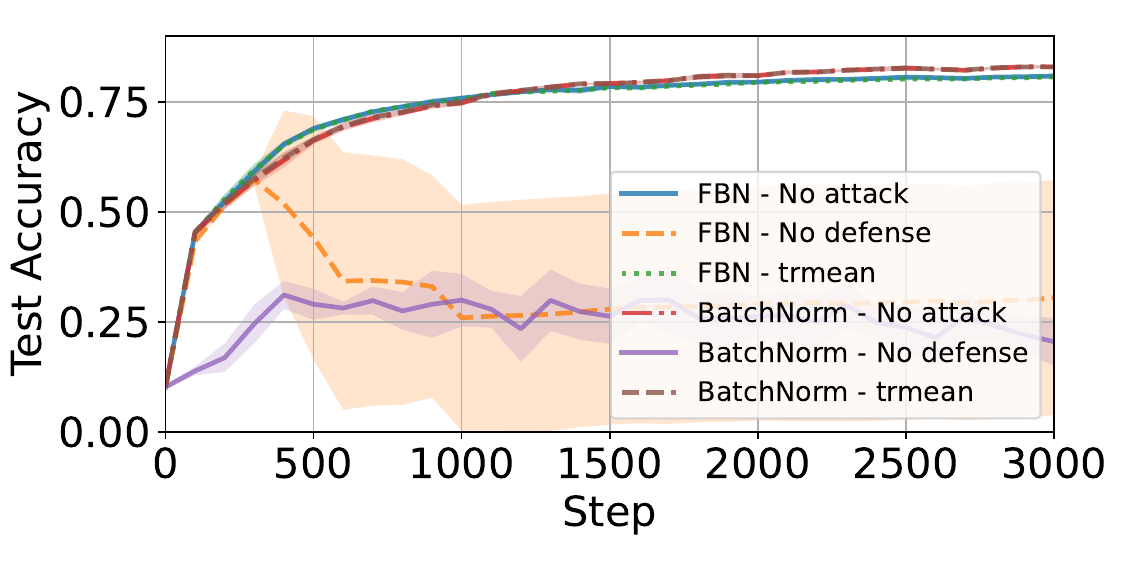}}
    \caption{Performance of \fbn{} vs. \bn{} on CIFAR-10 in two heterogeneity settings ($\alpha = 0.1$ and $\alpha = 1$), using the coordinate-wise trimmed mean~\cite{yin2018byzantine} aggregation.
    Among a total of 10 clients, we consider $f = 3$ adversarial clients executing the \textbf{ALIE}~\cite{baruch2019alittle}, \textbf{FOE}~\cite{xie2020fall}, and \textbf{SF}~\cite{allen2020byzantine} attacks.}
\label{fig_byz_app_dirichlet_trmean}
\end{figure*}

\end{document}